\documentclass{article}
\usepackage{jfrExamplee}
\usepackage{amsmath} % assumes amsmath package installed
\usepackage{amssymb}  % assumes amsmath package installed
\usepackage{threeparttable}
\usepackage{amsmath}
\usepackage{algorithm}
\usepackage{algorithmic}
\usepackage{amsfonts}
\usepackage{makecell}
\usepackage{multirow}
\usepackage{colortbl}
\usepackage{amsmath,amsfonts}
\usepackage{array}
\usepackage{textcomp}
\usepackage{stfloats}
\usepackage{url}
\usepackage{verbatim}
\usepackage{color}
\definecolor{Silver}{rgb}{0.752,0.752,0.752}
\definecolor{BlackHaze}{rgb}{0.913,0.921,0.917}
\definecolor{Manatee}{rgb}{0.58,0.58,0.584}
\usepackage{tabularray}
\usepackage{wasysym}
\usepackage{graphics} % for pdf, bitmapped graphics files
\usepackage{arydshln}
\usepackage{graphicx}
\usepackage{multirow}
\usepackage{pgf}

\usepackage{amssymb}  % assumes amsmath package installed
\usepackage{threeparttable}
\usepackage{booktabs}
\usepackage{hhline}
\usepackage{soul,color}
\soulregister\cite7
\soulregister\ref7
\soulregister\pageref7
\usepackage{color, colortbl}
	
\definecolor{LightCyan}{rgb}{0.88,1,1}
\usepackage{tikz}
\usepackage{amsfonts}
\usetikzlibrary{arrows}
\usetikzlibrary{shapes}
\usetikzlibrary{mindmap, shadows, arrows.meta,patterns,decorations.pathmorphing,decorations.markings,
                chains, shapes,backgrounds, shapes.multipart,
                positioning, shapes.geometric, arrows, positioning, calc, matrix}
                
\tikzset{
  basic box/.style={
    shape=rectangle, rounded corners, align=center,
    draw=#1, fill=#1!25},
  header node/.style={
    Minimum Width=header nodes,
    font=\strut\Large\ttfamily,
    text depth=+0pt,
    fill=white, draw},
  header/.style={%
    inner ysep=+1.5em,
    append after command={
      \pgfextra{\let\TikZlastnode\tikzlastnode}
      node [header node] (header-\TikZlastnode) at (\TikZlastnode.north) {#1}
      node [span=(\TikZlastnode)(header-\TikZlastnode)] at (fit bounding box) (h-\TikZlastnode) {}
    }
  },
  hv/.style={to path={-|(\tikztotarget)\tikztonodes}},
  vh/.style={to path={|-(\tikztotarget)\tikztonodes}},
  fat blue line/.style={ultra thick, blue}
}

\usepackage{nomencl}
\makenomenclature

\usepackage{color}
\definecolor{applegreen}{rgb}{0.55, 0.71, 0.0}

\usepackage[]{xcolor}
\definecolor{orange}{RGB}{255, 128, 0}
\definecolor{purple}{RGB}{102, 0, 204}
\definecolor{blue}{RGB}{0, 51, 102}
\definecolor{reds}{RGB}{102, 0,51}

\usepackage{todonotes}
\definecolor{myBlue}{RGB}{0, 0, 255}

\definecolor{MineShaft}{rgb}{0.121,0.121,0.121}
\usepackage{tabularray}

\title{Resilient Timed Elastic Band Planner for Collision-Free Navigation in Unknown Environments}

\author{Geesara Kulathunga$^{1}$, Abdurrahman Yilmaz$^{1}$, Zhuoling Huang$^{1}$, \\  Ibrahim Hroob$^{1}$,
Hariharan Arunachalam$^{1}$, Leonardo Guevara$^{1}$, Alexandr Klimchik$^{1}$, Grzegorz Cielniak$^{1}$, \\ and Marc Hanheide$^{1}$% <-this % stops a space
\thanks{This work was supported by the Innovate UK-funded project Agri-OpenCore [grant number 10041179].}% <-this % stops a space
\thanks{$^{1}$Lincoln Centre for Autonomous Systems, University of Lincoln, Lincoln, UK.
{\small gkulathunga@lincoln.ac.uk, ayilmaz@lincoln.ac.uk, zhuang@lincoln.ac.uk, ihroob@lincoln.ac.uk, 26760953@students.lincoln.ac.uk, lguevara@lincoln.ac.uk, aklimchik@lincoln.ac.uk, gcielniak@lincoln.ac.uk,  mhanheide@lincoln.ac.uk}}%
}
\begin{document}

\maketitle

\begin{abstract}
In autonomous navigation, trajectory replanning, refinement, and control command generation are essential for effective motion planning. This paper presents a resilient approach to trajectory replanning addressing scenarios where the initial planner's solution becomes infeasible. The proposed method incorporates a hybrid A* algorithm to generate feasible trajectories when the primary planner fails and applies a soft constraints-based smoothing technique to refine these trajectories, ensuring continuity, obstacle avoidance, and kinematic feasibility. Obstacle constraints are modelled using a dynamic Voronoi map to improve navigation through narrow passages. This approach enhances the consistency of trajectory planning, speeds up convergence, and meets real-time computational requirements. In environments with around 30\% or higher obstacle density, the ratio of free space before and after placing new obstacles, the Resilient Timed Elastic Band (RTEB) planner achieves approximately 20\% reduction in traverse distance, traverse time, and control effort compared to the Timed Elastic Band (TEB) planner and Nonlinear Model Predictive Control (NMPC) planner. These improvements demonstrate the RTEB planner's potential for application in field robotics, particularly in agricultural and industrial environments, where navigating unstructured terrain is crucial for ensuring efficiency and operational resilience.

\end{abstract}

\section{Introduction}

Motion planning is essential for autonomous navigation, allowing robots to move through complex environments. This process involves determining optimal trajectories that robots can follow while ensuring safety and efficiency. With advancements in robotics and artificial intelligence, there has been a shift towards online trajectory replanning, refining, and control command generation, crucial components methods that allow robots to adapt to real-time changes in their surroundings, incorporating feedback from their sensors~\cite{kulathunga2022optimization}. Unlike offline planning, which relies on pre-defined paths, online methods facilitate dynamic adjustments, making them essential for applications in environments with unpredictable obstacles. However, achieving the balance between computational efficiency and the generation of kinematically and dynamically feasible trajectories remains a significant challenge in the field. Researchers are continuously exploring innovative strategies to enhance motion planning algorithms, focusing on real-time optimisation and the integration of various planning components to improve the overall robustness and adaptability of autonomous systems~\cite{xiao2022motion}. However, formulating motion planning that ensures kinematic and dynamic feasibility while generating smooth control commands is computationally demanding~\cite{meng2024real}. 

The challenge in motion planning for autonomous robots stems from the need to integrate global and local planning components effectively. While global planning generates an initial path, it often overlooks the kinematic feasibility~\cite{allozi2022feasibility}, resulting in trajectories that the robot cannot physically follow. This limitation requires local planners to refine these paths, ensuring that the control commands respect both dynamic and kinematic constraints. Additionally, local planners must navigate a complex environment filled with both static obstacles, such as walls and trees, and dynamic ones like moving vehicles and pedestrians. This complexity demands rapid decision-making based on potentially noisy sensor data, making real-time optimisation a critical yet challenging task. Furthermore, local planners must balance safety and efficiency, optimising paths in real-time while accounting for the various constraints imposed by the environment~\cite{nair2024predictive}. 

This research investigates resilient trajectory replanning in autonomous navigation frameworks. In general, many navigation frameworks comprise two primary planning modules: global and local planners. Resilient trajectory replanning is particularly essential for local planners when the planning visibility range is constrained, as it enables the system to adapt to environmental changes and prevents the planner from becoming infeasible. 

The \textbf{contributions of this paper} are as follows:

% \begin{itemize}
    \textbf{Resilient trajectory generation}: A hybrid A* algorithm with a novel heuristic cost is proposed to generate such trajectories to handle cases where the TEB planner produces infeasible solutions. In such instances, the hybrid A* generates a new kinematically feasible trajectory, which is then used to reinitialise the TEB planner. This approach accelerates the convergence of the TEB planner and improves the consistency of trajectory planning, particularly in cluttered environments. It provides smoother control commands and reduces sudden directional changes, focusing on car-like vehicles. This is especially relevant for field applications where differential drive may not be feasible or desirable.
    
    \textbf{Resilient trajectory refinement procedure}: This procedure is framed as a soft constraint-based optimization to refine trajectories generated by the proposed hybrid A* algorithm. The refinement process incorporates considerations for trajectory continuity, obstacle avoidance, and kinematic feasibility. Obstacle constraints are defined using dynamic Voronoi maps, enabling smooth navigation through the constraint environments and reducing the need for replanning. As a result, the refined trajectories are both feasible and optimised for safety and efficiency in confined spaces.

    \textbf{Evaluating Resilient Planning in Obstacle-Constrained Scenarios}: a comprehensive evaluation of the proposed approach through real-world experiments that validate various components of the proposed approach under diverse obstacle constraints.

The primary practical advantage of RTEB lies in its recovery behaviour, which is triggered mainly in obstacle-dense environments, thereby reducing computational time. This approach leverages the strengths of the TEB planner along with the proposed recovery strategy, all while maintaining a moderate computational load.

\section{Related work}

Trajectory planning has evolved significantly, with various approaches developed to address the challenges posed by real-time, kinematic, and temporal constraints. Early methods like the Elastic Band (EB) approach~\cite{quinlan1995real} were initially introduced as a novel method that deforms the generated path while preserving the vehicle's distance from obstacles by defi
ning obstacle cost as a set of artificial forces. However, traditional trajectory planning methods generally do not incorporate kinematics and temporal constraints. Subsequently, In \cite{delsart2008reactive}, they proposed a reactive trajectory deformation method that, instead of path deformation, includes temporal constraints to predict the future behaviours of obstacles. Given the computational demands of the methods mentioned, various sampling-based techniques like DWA~\cite{fox1997dynamic} and MPPI~\cite{kim2022smooth} are well-suited for real-time trajectory planning in holonomic robots. However, sampling-based methods are not well-suited for trajectory planning in non-holonomic robots, such as car-like robots. 

To meet the temporal, real-time, and kinematic constraints, the TEB~\cite{rosmann2015timed} planner was developed. This planner mimics a predictive controller behaviour through variable-length receding horizon trajectory planning. The trajectory planning problem is formulated as a hyper-graph~\cite{rosmann2013efficient} and addressed as a sparse optimisation problem using the g2o~\cite{kummerle2011g} to ensure real-time computational efficiency. The Levenberg-Marquardt (LM) method~\cite{ranganathan2004levenberg} is utilised for solving the trajectory planning problem due to its robustness and efficiency. Since g2o implements a sparse variant of LM, where nonlinear constraint terms influence only a subset of parameters, it efficiently balances the underlying Hessian calculation, ensuring real-time performance with minimal constraint settings. 

Conversely, Model predictive control (MPC)-based trajectory planning~\cite{qie2022improved} for autonomous ground vehicles (AUV) has gained popularity over the past decade, owing to advances in computational embedded systems and control schemes for robotics. MPC can be formulated in two ways: convex and non-convex~\cite{kulathunga2023survey}. When system dynamics and constraints are non-linear or non-convex, it is necessary to use NMPC~\cite{kulathunga2022trajectory}. This may involve linearisation, applying Sequential Quadratic Programming (SQP) variations~\cite{torrisi2016variant}, or directly solving the non-linear formulation. Solving MPC non-linearly typically involves multiple shooting and direct collocation methods~\cite{kulathunga2022trajectory}. Many MPCs aim to minimise a cost function that penalises a combination of control error and other specified costs~\cite{rosmann2015timed} subject to provided hard and soft constraints. Time-optimal, point-to-point transitions~\cite{van2011model} are generally not well-developed for real-time applications. In~\cite{rosmann2015timed}, TEB was extended into MPC to achieve time-optimal point-to-point transitions. While MPC usually has a fixed receding horizon length, this work formulated MPC using TEB, retaining a discrete variable time interval between consecutive optimisation parameters. The MPC formulation becomes highly non-linear in cluttered environments, leading to high computational demands. Unless system dynamics are highly non-linear, TEB is an optimal choice as a local planner for autonomous navigation. 

The TEB planner is not specifically designed for horizon-based planning methods like MPC. Instead, it utilises a variable horizon length approach, which focuses on point-to-point trajectory planning~\cite{stefanini2024efficient}. Intermediate goals are derived from a global path, enabling a quasi-path-following strategy through waypoint navigation. However, the TEB planner's computation time increases as the number of feasible points grows, due to the need to verify kinematic feasibility\cite{wullt2024model}. While this approach converges more quickly in obstacle-free environments, it takes longer to reach a final consistent feasible trajectory in environments with many obstacles. In such cases, waypoint navigation with many feasible points is less practical, as intermediate poses may end up within obstacle zones, leading to significantly higher computational costs. Moreover, if the global planner does not specify the orientations of the poses, the TEB planner defaults to a forward-oriented planning approach, aligning each pose's orientation with the direction of the next pose in the path. This is less efficient when the robot can navigate in both forward and backward directions\cite{huajian2024sample}. The following timeline illustrates the evaluation of the timed elastic band planner and its variations in the recent past. 

\begin{tikzpicture}[
node distance = 1mm and 3mm,
  start chain = A going below,
   dot/.style = {circle, draw=white, very thick, fill=gray!60, minimum size=3mm},
   box/.style = {rectangle, text width=140mm, inner xsep=4mm, inner ysep=1mm, font=\normalsize		\linespread{0.84}\selectfont, on chain},]
    \begin{scope}[every node/.append style={box}]
    \node {Social Elastic Band: predictive path optimisation for socially aware robot navigation\cite{perez2024social}};
    \node {A nonlinear proportional function of tracking error recalibrates the robot control quantity~\cite{10135320}};
    \node {DTEB Planner: Trajectory Sharing and Collision Prediction for Multi-Robot Systems \cite{chung2022distributed}};
    \node {Real-time motion planning utilising nonlinear model predictive control in conjunction with non-Euclidean rotation groups~\cite{9654872}};
    \node { Dynamic obstacle-aware model predictive control for collaborative manipulators~\cite{kramer2020model}};
    \node {Time-optimal nonlinear model predictive control~\cite{rosmann2019time}};
    \node {Timed Elastic Bands for efficient online motion planning of car-like robots~\cite{8206458}};
    \node {TEB for time-optimal point-to-point nonlinear model predictive control~\cite{rosmann2015timed}};
    \node {“g2o” for solving “timed elastic band” problems~\cite{6698833}};
    \node {The "timed elastic band" problem is expressed within a framework of weighted multi-objective optimisation~\cite{6309484}};
    \end{scope}
\draw[very thick, gray, {Triangle[length=4pt)]}-{Circle[length=3pt]},
      shorten <=-3mm, shorten >=-3mm]           % <--- here is adjusted additional arrow's 
    (A-1.north west) -- (A-10.south west);
\foreach \i [ count=\j] in {2024, 2023, 2022, 2021, 2020, 2019, 2017, 2015, 2013, 2012}
    \node[dot,label=left:\i] at (A-\j.west) {};
    \end{tikzpicture}

Conversely, a substantial body of prior research has introduced learning-based approaches for trajectory planning, with many utilising simulated environments to collect data for training purposes \cite{kadian2020sim2real, chaplot2020semantic, tan2019learning, chen2019learning}. Subsequently, various methods have employed real-world experiments to learn policies that generate control commands \cite{fang2023generalization, khazatsky2021can, shah2021rapid}. These policies are primarily trained using reinforcement learning techniques. However, due to the inherent stochastic nature of these algorithms, they often struggle to generalise effectively to previously unseen environments \cite{kadian2020sim2real, gervet2023navigating}. To address these limitations, recent work has proposed the use of hierarchical reinforcement learning and generative models combined with high-level global planning \cite{faust2018prm, li2020hrl4in, shah2023vint}. While these approaches offer improvements, they also introduce additional computational complexity, rendering them unsuitable for real-time trajectory planning tasks. As a result, learning-based trajectory planning remains in its early stages and is not yet mature enough for practical real-world applications. In contrast, analytical methods continue to outperform these learning-based approaches, providing deterministic estimations of control commands, which offer higher confidence and reliability for deployment in real-world scenarios. 

In this work, the ROS2 navigation stack was selected to validate the proposed RTEB planner. Commonly used planners within ROS2 include TEB~\cite{rosmann2017kinodynamic}, NMPC~\cite{rosmann2021online}, the Dynamic Window Approach (DWA), and Model Predictive Path Integral (MPPI), among others~\cite{houshyari2022new, yucel2021fuzzy}. Among the available options, TEB and NMPC are the most optimal for generating control commands, particularly for car-like robots, which is one of the main objectives, each utilising distinct techniques~\cite{rosmann2015timed, rosmann2021online}. Table~\ref{tab:pros_cons} lists the characteristics, pros, and cons of the aforementioned planners, including the proposed approach.

\begin{table}[ht!]
\label{tab:pros_cons}
\centering
\caption{The characteristics, pros, and cons of RTEB, TEB, NMPC, and MPPI trajectory planning methods.}
\begin{tabular}{|l|l|l|} 
\hline
\textbf{Method} & \textbf{Pros}                                                                                                                                                                                                                              & \textbf{Cons}                                                                                                                                                                             \\ 
\hline
\textbf{RTEB}   & \begin{tabular}[c]{@{}l@{}}- High success rate in complex environments \\- Moderate computation time  \\- Bidirectional orientation flexibility (Hybrid A*)\end{tabular}                                                        & \begin{tabular}[c]{@{}l@{}}- Higher computational cost \\~ compared to TEB when activating \\~ resilient planning \\- More complex implementation\end{tabular}                                                                   \\ 
\hline
\textbf{TEB}    & \begin{tabular}[c]{@{}l@{}}- Moderate success rate \\- Efficient in environments with moderate \\~ complexity\cite{rahmani2020teb} \\- Low computation time\cite{li2019neural} \end{tabular}                                                                                        & \begin{tabular}[c]{@{}l@{}}- Lower success rate in challenging \\~ ~environments~\cite{wu2021improved} \\- Forward-oriented, limited orientation \\~ flexibility \\- Poor handling of narrow \\ gaps~\cite{wu2021improved}\end{tabular}  \\ 
\hline
\textbf{NMPC}   & \begin{tabular}[c]{@{}l@{}}- High path efficiency~\cite{ge2023efficient} \\- Capable of handling dynamic and constrained\\~ systems\end{tabular}                                                                    & \begin{tabular}[c]{@{}l@{}}- Very high computational cost \\~\cite{astudillo2024rapid}\\ - Low efficient in complex environments \\- High computation time\\~\cite{astudillo2024rapid}\end{tabular}                                                                                        \\ 
\hline
\textbf{MPPI}   & \begin{tabular}[c]{@{}l@{}}- Good balance between computation time\\~ and path efficiency~\cite{kazim2024recent} \\- Due to its stochastic behaviour, \\~ performance depends on the implementation\end{tabular} & \begin{tabular}[c]{@{}l@{}}- Less success rate in complex \\environments~\cite{kazim2024recent} \\- Moderately high computational cost \\- Requires fine-tuning of parameters \\ ~\cite{kazim2024recent}\end{tabular}                                                                       \\
\hline
\end{tabular}
\end{table}

The primary drawback of the TEB and NMPC planners is that their computation time increases exponentially with the number of feasible points in TEB and the receding horizon length in NMPC, both of which require kinematic feasibility checks. In environments with fewer obstacles, such drawbacks can be alleviated by reducing the number of feasibility-checking points or shortening the planning horizon. However, substantial computational resources and time are required to generate feasible control commands. To tackle this issue, we introduced a resilient trajectory generation method followed by a smoothing technique for the TEB planner. Thus, the proposed approach is designed to regain functionality when the trajectory generated by the initial planner does not meet kinematic feasibility requirements to introduce resilient planning that enables recovery while managing moderate computational demands.

\section{Methodology}
\begin{figure}[h!t]
    \centering
    \includegraphics[width=0.7\linewidth]{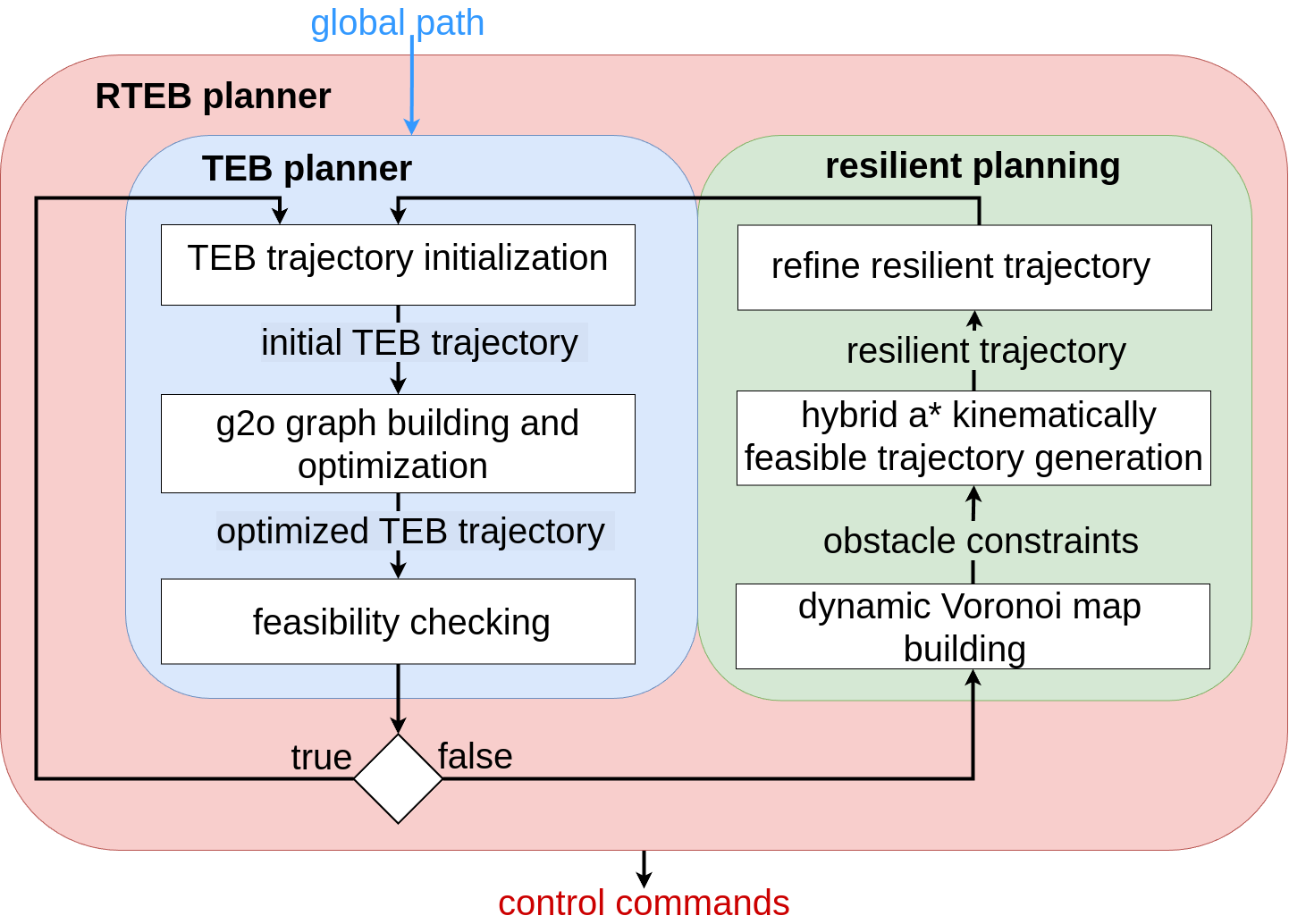}
    \caption{Resilient Timed Elastic Band (RTEB) Planner Architecture: The resilient planning module extends the standard TEB planner by introducing enhanced recovery capabilities.}
    \label{fig:rteb_concept}
\end{figure}

\begin{figure}
    \centering
    \includegraphics[width=1.0\linewidth]{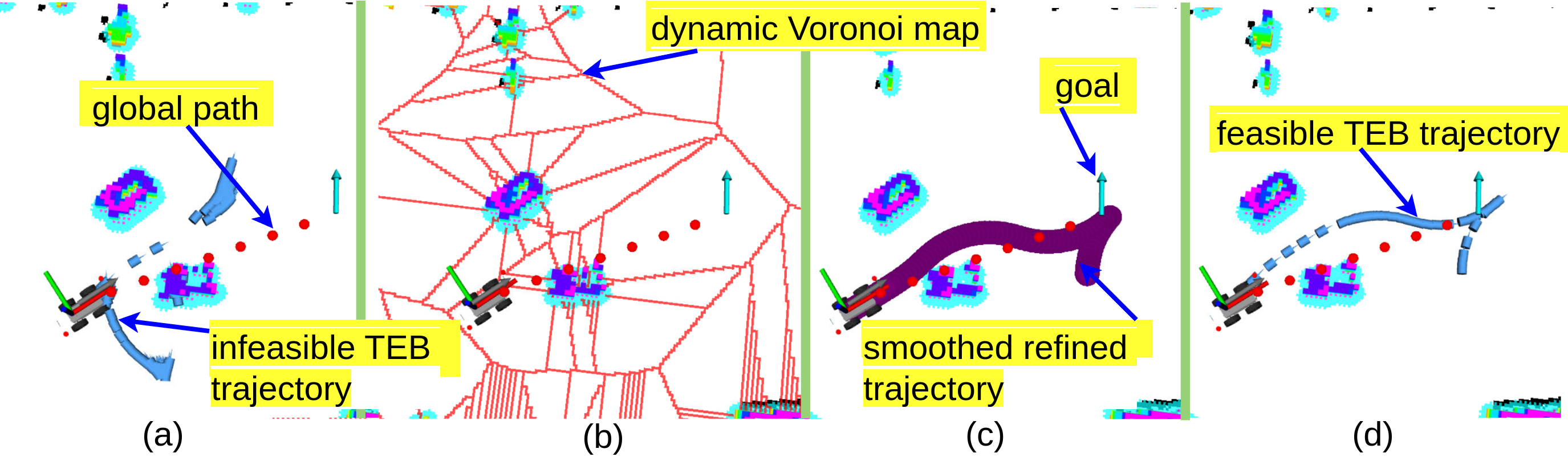}
    \caption{Trajectory planning using the RTEB planner -- (a) The TEB planner initially fails to find a solution. (b) The dynamic Voronoi graph-based Voronoi field in \eqref{eq:voronoi_field} aids in pushing the hybrid A* planned trajectory further away from obstacles. (c) Employing the proposed hybrid A* planner followed by smoothing yields a feasible trajectory. (d) Reinitialisation of the TEB planner according to the planned path by Hybrid A* and the newly planned feasible path. The global path was generated as a set of straight lines without considering obstacle avoidance to prevent any bias in the local planning.}
    \label{fig:recovery_planner_idea}
\end{figure}

Building upon the advancements and limitations identified in previous methods, the proposed RTEB planner introduces several key innovative resilient (recovery) plans (Fig.\ref{fig:rteb_concept}) to enhance the TEB planner~\cite{rosmann2017integrated}. First, it implements variable-length feasible global path generation, improving the efficiency of waypoint navigation. Second, during each planning iteration, the trajectory's kinematic feasibility is assessed. If deemed infeasible, a hybrid A* planner is employed, followed by a smoothing process to reinitialise the TEB planner's current poses and update its optimisation (see Fig.\ref{fig:recovery_planner_idea}). This approach allows for rapid convergence to a feasible trajectory, bypassing the TEB planner's slower and independent optimisation process. Additionally, the hybrid A* planner generates each intermediate pose with its orientation, independent of the global plan’s orientation, ensuring proper orientation adjustments when refining the TEB planner’s poses. This orientation estimation enables the robot to navigate efficiently in both forward and backward directions. Additionally, the RTEB supports precise goal alignment operations; for example, the robot's orientation must closely match the desired alignment at the specific target location.

The following subsections highlight the key advancements introduced in the RTEB planner. Section~\ref{sec:problem_formulaiton} provides a mathematical formulation of the RTEB planner. In Section~\ref{sec:hybrid_a}, a hybrid A* algorithm with a novel heuristic cost is integrated to generate kinematically feasible trajectories when TEB trajectory generation fails, enabling seamless reinitialization of the TEB planner. Section~\ref{sec:smoothing} discusses the trajectory refinement process, which pushes the path into free space to minimize the need for frequent replanning. Finally, Section~\ref{sec:goal_alignment} introduces an accurate goal alignment procedure to ensure precise robot orientation at the target goal.

\subsection{Resilient Timed Elastic Band planner: problem statement}\label{sec:problem_formulaiton}
The RTEB planner is formulated as a trajectory optimisation problem for a mobile robot, where each state \(\mathbf{x} = [x, y, \theta]^\top\) denotes the robot’s position \((x, y)\) and orientation \(\theta\). The goal is to gBuilding upon the advancements and limitations identified in previous methods, tnerate a collision-free, smooth, and kinematically feasible trajectory \(\mathcal{T} = \{\mathbf{x}_t\}_{t=0}^{T}\) over a finite planning horizon \(T\). Control inputs \(\mathbf{u} = [\phi, s]^\top\) consist of the steering angle \(\phi\) and motion arc length \(s\), constrained to ensure the vehicle’s manoeuvrability. To meet trajectory refinement objectives, we minimise a cost function \(J_{{total}}\), which combines weighted sub-costs: obstacle avoidance \(J_{{obs}}\), curvature smoothness \(J_{{cur}}\), and path efficiency \(J_{{path}}\). Obstacle avoidance leverages a dynamic Voronoi field, penalising proximity to obstacles, while curvature and path length costs promote smoothness and directness of the trajectory. Additionally, goal alignment is achieved by ensuring that the trajectory approaches an intermediate pose near the target, with specific orientation \(\theta_g\), to refine accuracy. This mathematical framework will guide the notation and conversion methods used in the subsequent sections of the paper.

\subsubsection{Hybrid A* with novel heuristic cost: recovery trajectory generation}\label{sec:hybrid_a}

Since we focus on car-like vehicles, the discrete-time state transition is expressed in the following way:

\[
\mathbf{x}_{t+1} = \mathbf{x}_t +
\begin{cases}
\begin{pmatrix}
k (\sin(\theta_t + \frac{s}{k}) - \sin(\theta_t)) \\
-k (\cos(\theta_t + \frac{s}{k}) - \cos(\theta_t)) \\
\frac{s}{k}
\end{pmatrix}, & \text{if } \phi \neq 0 \\[2em]
\begin{pmatrix}
s \cos(\theta_t) \\
s \sin(\theta_t) \\
0
\end{pmatrix}, & \text{if } \phi = 0
\end{cases}
\] where term $k = L/\tan(\phi)$  and robot wheelbase is denoted by $L$. To define the initialisation of state propagation in hybrid A* planning, let $\phi_{\max}$ be the maximum steering angle and the term $r_s$ be the resolution parameter for arc length division. Control input $\mathcal{U}$ set is defined for forward ($v_t \geq 0$) and backward motions  ($v_t < 0$) as follows: 

\[
\mathcal{U} = \left\{
\begin{aligned}
&\left\{(\delta, a) \mid -\delta_{\max} \leq \delta \leq \delta_{\max}, \; \delta = r_s \cdot \delta_{\max}, \; r_s \in \mathbb{R}, \right. \\
&\left. \quad \text{and} \quad a = r_a \cdot s_{max},  \; r_a \in \mathbb{R},\right. 
\left.  \begin{cases}
s_{max} \leq a \leq 2 \cdot s_{max}, \; \text{if } v_t \geq 0 \\
-2 \cdot s_{max} \leq a \leq -s_{max}, \; \text{if } v_t < 0
\end{cases}\right\}
\end{aligned}
\right.
\] 
% \begin{multline*}
%  \mathcal{U} = \{(\delta, a) \mid -\delta_{\max} \leq \delta \leq \delta_{\max}, \; \delta = r_s \cdot \delta_{\max}, \; r_s \in \mathbb{R} 
%  \\ 
% \quad \text{and} \quad 
% a = r_a \cdot s_{max}, \;  r_a \in \mathbb{R}\}
% \\
% \left\{\begin{matrix}
%  s_{max} \leq a \leq 2 \cdot s_{max}, \text{if } \; v_t \geq 0
% \\
% -2 \cdot s_{max} \leq a \leq -s_{max}, \text{if }\; v_t < 0
% \end{matrix}\right.   
% \end{multline*}
\noindent
where steering angle \(\delta\) is constrained to \([- \delta_{\max}, \delta_{\max}]\) and given by \(\delta = r_s \cdot \delta_{\max}\). For non-negative velocity \(v_t \geq 0\), the arc length \(a\) ranges from \(s_{\max}\) to \(2 \cdot s_{\max}\) and is scaled as \(a = r_a \cdot s_{\max}\), with \(r_a \in \mathbb{R}\). For negative velocity \(v_t < 0\), \(a\) ranges from \(-2 \cdot s_{\max}\) to \(-s_{\max}\), similarly scaled by \(s_{\max}\). When the velocity \(v_t\) is non-negative, the inputs include positive accelerations; otherwise, they account for negative accelerations. The parameters used in the experiments, \(r_s = 0.3\), \(r_a = 0.5\), and \(s_{{max}} = 1.0\), were determined through unbiased testing.

In the hybrid A* planner, the heuristic \(g^p_{{score}}\) and \(f^p_{{score}}\) are calculated as follows:
\[
\begin{aligned}
    g^p_{{score}} &= g^c_{{score}}  + \begin{cases}
    \lambda_f \cdot |a| & \text{if } \zeta > 0 \\
    \lambda_b \cdot |a| & \text{if } \zeta \leq 0
    \end{cases} + \lambda_s \cdot |\delta| \cdot |a|  + \lambda_{sc} \cdot |\delta - \phi_t|,  \zeta = \begin{cases}
   1 & \text{if } a > 0, \\
   -1 & \text{if } a \leq 0. 
   \end{cases} \\
   f^p_{{score}} &= g^p_{{score}}  + \lambda_{{heu}} \cdot \eta \cdot \| \mathbf{x}_{{t}} - \mathbf{x}_{{g}} \|_2
\end{aligned}
\] In this formulation, \(g^p_{{score}}\) represents the cost associated with transitioning from the current node to a specific node, where each node includes the robot's desired state and control inputs during graph search.

It incorporates several components: \(g^c_{{score}}\) is the base cost from the current node, while \(\lambda_f\) and \(\lambda_b\) are penalties for forward and backward arc length changes, respectively. The term \(\lambda_s \cdot |\delta| \cdot |a|\) accounts for the penalty associated with steering, and \(\lambda_{sc} \cdot |\delta - \phi_t|\) penalises changes in the steering angle.  The cost to goal \(f^p_{{score}}\) is obtained by adding a heuristic component to \(g^p_{{score}}\). This heuristic component, \(\lambda_{{heu}} \cdot \eta \cdot \| \mathbf{x}_{{t}} - \mathbf{x}_{{g}} \|_2\), uses the Euclidean distance between the current state \(\mathbf{x}_{{t}}\) and the goal state \(\mathbf{x}_{{g}}\), scaled by the heuristic weight \(\lambda_{{heu}}\) and a tie-breaking \cite{horne2005dynamic} factor $\eta$. The parameters that were used in experiments are $\lambda_f=\lambda_b=1.0, \lambda_s=0.5, \lambda_{sc} = 0.01, \lambda_{heu}=5.0, \eta=1.0001$, were estimated empirical testing.  These parameters are integrated into the RTEB planner to balance smoothness, acceleration, and speed constraints, ensuring the generation of feasible and dynamically consistent trajectories that adhere to the robot's physical limitations.

\subsubsection{Trajectory refinement: enhancing free space navigation to minimise replanning}\label{sec:smoothing}

The hybrid A* planned trajectory will be further refined using the proposed smoothing technique as outlined below. Smoothing helps push the planned trajectory away from obstacles that help to RTEB planner generate consistent control commands. The overall objective function for smoothing is expressed as:
\begin{equation}
    \begin{aligned}
        J_{{total}} &= \lambda_{{obs}} J_{{obs}} + \lambda_{{cur}} J_{{cur}} + \lambda_{{path}} J_{{path}},
    \end{aligned}
\end{equation} where \(J_{{obs}}\), \(J_{{cur}}\), and \(J_{{path}}\) are the costs associated with obstacles, path curvature, and path length improvement, respectively. The weights \(\lambda_{{obs}} = 0.5\), \(\lambda_{{cur}} = 0.3\), and \(\lambda_{{path}} = 0.2\) were set based on empirical testing, prioritising obstacle cost over the other costs.

The obstacle cost, $J_{{obs}}$, is evaluated using the Voronoi field approach \cite{lau2010improved}, which improves upon the Artificial Potential Fields (APF) method. The APF method can create high-potential areas near narrow passages that obstruct robot movement, whereas the Voronoi field adjusts the potential based on the configuration space's geometry, facilitating better navigation through tight spaces \cite{dolgov2008practical}. The obstacle cost is defined by:
\begin{equation}
    \begin{aligned}
        J_{{obs}} = \sum_{p=0}^{Q} F_v(x_p, y_p),
    \end{aligned}
\end{equation}
where \(Q\) is the number of points along the planned trajectory. The Voronoi field value \( F_v(x_p, y_p) \) is expressed as:
% \begin{equation}
% \begin{aligned}
%     F_v(x_p, y_p)
%        = \left[1- \left(\frac{d_{obs}(x_p, y_p)}{d_{obs}^{max}}\right)^2\right]
%        \cdot  \left(\frac{\lambda_v}{\lambda_v + d_{obs}(x_p, y_p)}\right) 
%        \cdot \left(\frac{d_{vor}(x_p, y_p)}{d_{obs}(x_p, y_p) + d_{vor}(x_p, y_p)}\right)
%     \end{aligned}
% \end{equation}

\begin{equation}\label{eq:voronoi_field}
    F_v(x_p, y_p) = A(x_p, y_p) \cdot S(x_p, y_p) \cdot R(x_p, y_p)
\end{equation}

\noindent where:
\begin{align*}
    A(x_p, y_p) &= 1 - \left(\frac{d_{obs}(x_p, y_p)}{d_{obs}^{{max}}}\right)^2,
    S(x_p, y_p) &= \frac{\lambda_v}{\lambda_v + d_{obs}(x_p, y_p)}, 
    R(x_p, y_p) &= \frac{d_{vor}(x_p, y_p)}{d_{obs}(x_p, y_p) + d_{vor}(x_p, y_p)}.
\end{align*}

\noindent
Here, \( \lambda_v > 0 \in \mathbb{R} \) governs the decay of influence for the Voronoi field \cite{lau2013efficient}, and \( d_{obs}^{max} > 0 \in \mathbb{R} \) is the maximum allowable distance between the robot and closest obstacle pose. The function \( d_{obs}(\cdot) \) indicates the distance to the nearest obstacle, while \( d_{vor}(\cdot) \) indicates the distance to the nearest Voronoi diagram edge.

% \begin{equation}
%     \begin{aligned}
%        F_v(x_p, y_p)
%        =  \left(1 - \frac{d_{obs}(x_p, y_p)}{d_{obs}^{max}}\right)^2 \\
%        \cdot  \left(\frac{\lambda_v}{\lambda_v + d_{obs}(x_p, y_p)}\right) 
%        \cdot \left(\frac{d_{vor}(x_p, y_p)}{d_{obs}(x_p, y_p) + d_{vor}(x_p, y_p)}\right) \\
%     \end{aligned}    
% \end{equation} 

% \noindent 
% where \( \lambda_v > 0 \in \mathbb{R} \) governs the decay of influence for the Voronoi field \cite{lau2013efficient}, and \( d_{obs}^{max} > 0 \in \mathbb{R} \) is the maximum allowable distance. The function \( d_{obs}(\cdot) \) indicates the distance to the nearest obstacle, while \( d_{vor}(\cdot) \) indicates the distance to the nearest Voronoi diagram edge.

The curvature penalty \(J_{{cur}}\) measures how much the instantaneous curvature \(k_p\) deviates from a predefined maximum curvature \(k_{{max}}\). This penalty is computed by summing the squared differences between \(k_p\) and \(k_{{max}}\). Here, \(k_{{max}}\) is the reciprocal of the minimum radius of curvature \(\rho_{{min}}\), and it applies uniformly regardless of the curvature's sign. The instantaneous curvature \(k_p\) is derived from the ratio of the angle change \(\delta \theta_p\) to the positional change \(\delta x_p\).
\begin{equation}\label{eq:cost_k}
    \begin{aligned}
        J_k =& \sum_{p=0}^{Q-1} \left(k_p - 
        \begin{Bmatrix}
            1  & \text{if } k_p > 0 \\
            0  & \text{if } k_p = 0 \\
           -1  & \text{if } k_p < 0
        \end{Bmatrix}
        \cdot k_{{max}}\right)^2, \\
        k_p =& \Big(\text{atan2}(\Delta y_{p+1}, \Delta x_{p+1}) - \text{atan2}(\Delta y_p, \Delta x_p)\Big) / \Delta \mathbf{x}_p,
    \end{aligned}
\end{equation} where \(\Delta x_p = x_p - x_{p+1}\) and \(\Delta y_p = y_p - y_{p+1}\) represent the differences in x and y coordinates between consecutive waypoints at the \(p\)-th index. The cost function for improving path length is defined by:
\begin{equation}
J_{{path}} = \sum_{p=0}^{Q-1} \left\| (\Delta x_{p+1}, \Delta y_{p+1}) - (\Delta x_p, \Delta y_p) \right\|^2
\end{equation} 

The \(J_{{path}}\) measures the deviation in path length by penalising changes in displacement between successive waypoints, encouraging a smoother trajectory with fewer abrupt changes in direction. This trajectory refinement process not only enhances trajectory feasibility by minimizing abrupt changes but also ensures safer navigation through complex environments, significantly improving the RTEB planner's robustness and consistency.

\subsubsection{Precise goal alignment}\label{sec:goal_alignment}

\begin{figure}[ht!]
    \centering
    \includegraphics[width=1.0\linewidth]{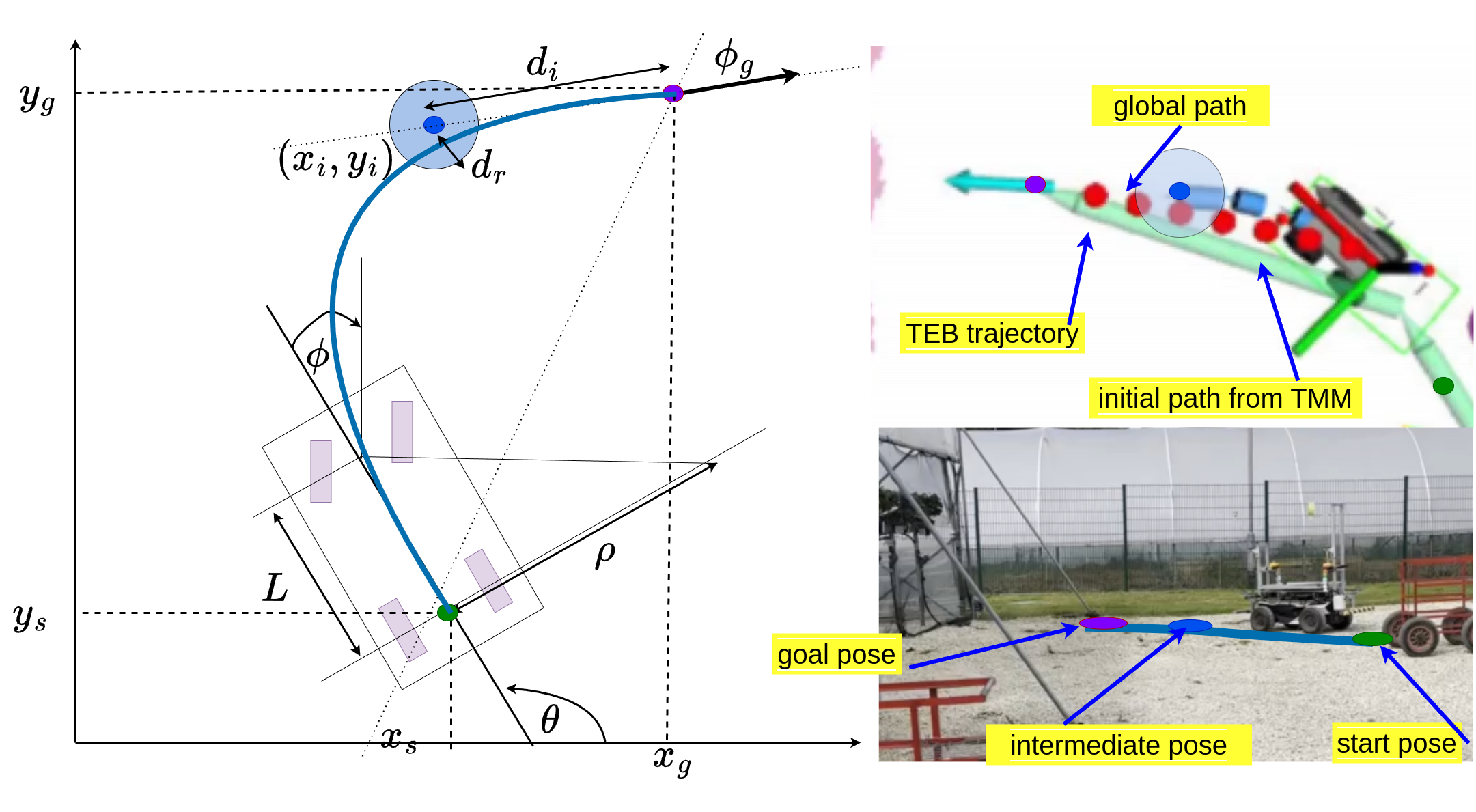}
    \caption{The goal alignment behaviour of the RTEB planner is particularly crucial at the start of in-row navigation. This behaviour ensures that the vehicle aligns its orientation with the desired direction, facilitating a smoother and more accurate path following within the row.}
    \label{fig:goal_alignment_idea}
\end{figure}

When the RTEB planner gets closer to the final goal, the proposed precise goal alignment procedure is activated as illustrated in Fig. \ref{fig:goal_alignment_idea}. Following this, the RTEB planner calculates an intermediate pose \((x_i, y_i, \phi_g)\) that is displaced by a distance \(d_i \in C, e.g., 1.0 m,\) from the goal pose \((x_g, y_g, \phi_g)\). The intermediate pose coordinates are determined by:
\[ x_i = x_g - d_i \cos(\phi_g), y_i = y_g - d_i \sin(\phi_g), \]
where \((x_i, y_i)\) represents the intermediate position, and \(\phi_g\) denotes the goal orientation. The robot will start moving towards the goal pose as long as the distance between its current pose \((x_s, y_s, \phi)\) and the intermediate pose \((x_i, y_i)\) is within a threshold \(d_r \in C, i.e., 0.1 m\): $\sqrt{(x_s - x_i)^2 + (y_s - y_i)^2} \leq d_r$. This method ensures that the robot is aligned with the desired goal orientation before reaching the goal pose, thereby enhancing the precision of the final orientation at the goal.

By incorporating this precise alignment method, the RTEB planner ensures not only accurate goal-reaching behavior but also smooth and consistent orientation adjustments. This capability significantly improves the reliability and effectiveness of the planner in scenarios requiring high-precision navigation.

\subsection{Integration of the RTEB with the ROS2 navigation stack}

To validate the performance of the RTEB planner, we integrated it with the ROS2 navigation stack. The ROS2 navigation stack splits planning into two stages: global and local. In our setup, the proposed RTEB planner was employed as the local planner, while a custom global planner was developed to simplify the process. This global planner bypasses obstacle avoidance, generating a straight-line path between intermediate poses from the robot's current position to the goal. Its primary function is to provide a consistent global path for the RTEB planner, whose evaluation takes place during the local planning stage. 

\subsubsection{Global planning with topological map manager}

\begin{figure*}[ht!]
    \centering
    % \vspace{0.3cm}
    \includegraphics[width=1.0\linewidth]{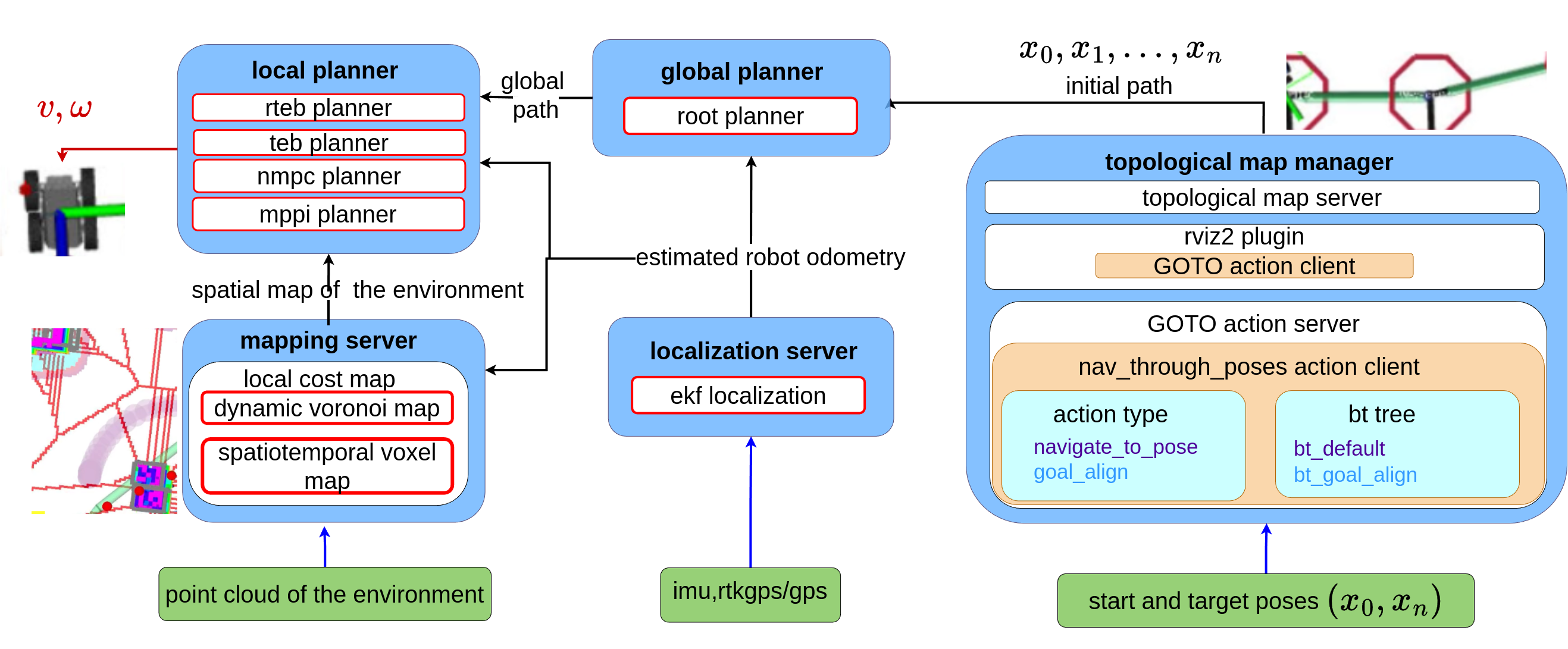}
    \caption{The proposed navigation stack is built on the ROS2 navigation stack. It has specific ROS2-compatible plugins for global and local planning as well as local and global mapping. Topological map manager helps to generate an initial high-level root plan that subscribes by  \textit{nav\_through\_poses\_action\_server} along with a specific behaviour tree that depends on the action type that topological map manager provides.}
    \label{fig:agri_core}
\end{figure*}

For localisation, we utilised RTK-GPS, and the mapping is handled through a spatiotemporal voxel mapper, as detailed in the following subsections. To generate global planning, we developed a Topological Map Manager (TMM)\footnote[1]{\url{https://github.com/GPrathap/topological_navigation} \label{toponav}} (see Fig. \ref{fig:agri_core}), which functions as a high-level planning utility that guides robot behaviour across the provided environment. The TMM represents a high-level traversability graph as a set of nodes and edges, where each node attaches to a specific navigation action within the environment. Initially, the TMM performs root planning by navigating along the edges of the topological map, based on the robot's current and target poses. Subsequently, the TMM generates an action message, which is processed by the \textit{nav\_through\_poses\_action\_server} \cite{ghzouli2023behavior} ROS2 navigation behaviour server. The TMM's backend interfaces with the ROS2 navigation behaviour server, relaying responses to the TMM's GOTO action server (Fig. \ref{fig:agri_core}), which serves as an intermediary between the ROS2 navigation behaviour server and front-end applications such as RViz. The ROS2 global planner then receives the initial path from the TMM and regenerates the global path as a series of straight lines. If the robot’s pose approaches the path's edge, for example within 2 m (a configurable threshold), the planner will skip that close edge and connect to the subsequent one. The regenerated global path is shown as red dashed lines in Fig. \ref{fig:recovery_planner_idea}. This global planning is performed without obstacle avoidance, primarily to ensure that the local planner receives the same global path, preventing any bias in its performance caused by the global plan. However, the proposed framework can also accommodate other global planning techniques.

\subsection{Localisation and mapping}
For localisation, we utilise RTK-GPS, which is particularly well-suited for outdoor navigation scenarios. On the other hand, for mapping, we implement a spatiotemporal voxel mapper, serving as the local cost map \cite{1729881420910530}. Given the project’s focus on utilising low-cost sensors, we employ Livox LiDAR \cite{lin2020loam}, which, while cost-effective, tends to produce a significant amount of false positive data. Additionally, outdoor environments, in general, are characterised by uneven surfaces, further complicating the accurate interpretation of sensor data.  

Consequently, it is crucial to address the challenge of distinguishing between ground and non-ground points in the input point clouds before integrating the data into the spatiotemporal voxel mapper. For this reason, the rough ground itself may sometimes be mistakenly perceived as an obstacle. In contrast, at other times, false positive measurements caused by the robot's vertical oscillatory movements due to ground roughness can lead to incorrect detections. This preprocessing step is essential for mitigating the impact of false positives and ensuring the reliability of the mapping process. For a given set of point clouds, such as those obtained from LiDAR or depth cameras, the mapping utility processes the data to estimate the optimal spatial-temporal voxel clusters representing the environment. The processing workflow begins with ground removal, a critical step to isolate the relevant points from those representing the ground surface. For this task, Patchwork++ \cite{lim2021patchwork} is utilised, effectively removing the ground points from the initial point clouds. 

\section{Experimental procedure and results}
In our study, we aimed to rigorously evaluate the performance of our proposed RTEB planner, specifically against two established planners, TEB and NMPC. These planners were implemented as local planners within the ROS2 navigation stack to support a consistent comparison framework. Our experiments were structured into two primary evaluations. The first experiment focused on goal alignment, comparing RTEB and TEB across various scenarios to assess each planner's accuracy in maintaining goal-oriented navigation. The second experiment extended this evaluation to real-world and simulated environments with diverse obstacle densities, enabling a detailed comparison of RTEB's robustness and adaptability compared to TEB and NMPC.

\subsection{Experimental environment}
The experimental environment for both simulation and real-world trials was based on a section of a strawberry farm located at the University of Lincoln, UK. As shown in Fig. \ref{fig:hardware_setup}, this environment provided a realistic and challenging setting for testing the proposed approach. To maintain consistency, the same environment was replicated in the Gazebo simulator, allowing for extensive testing and refinement of the navigation strategies in a controlled virtual setting. This approach facilitated iterative development, reducing the need for frequent field trials while ensuring the software's readiness for real-world deployment. All simulated experiments took place in Gazebo, where the farm environment was modelled to match the physical setup closely. Using this simulated environment allowed for a safe and cost-effective method to try different navigation algorithms, especially under various controlled conditions that would be difficult to replicate outdoors. By fine-tuning the system within the simulator, we were able to ensure a higher level of robustness before conducting tests in the actual field.

\subsection{Real-world platform: hardware and software stack overview}
\begin{figure}[h!t]
    \centering
    \includegraphics[width=1.0\linewidth]{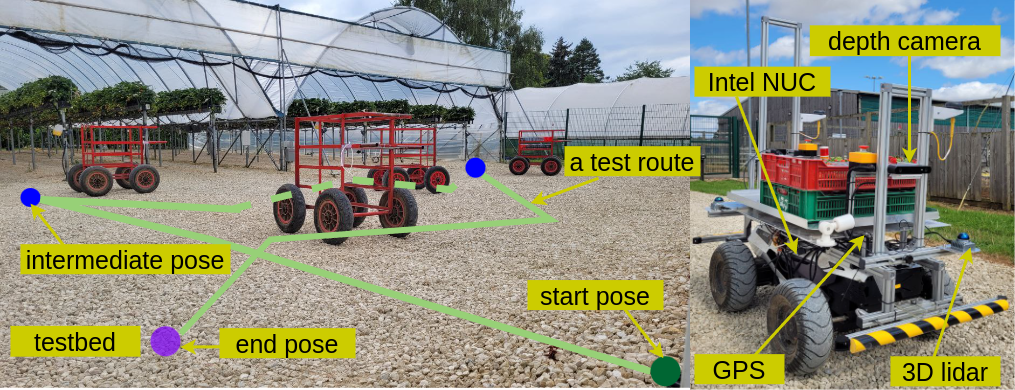}
    \caption{Real-world testing environment and robot equipped with sensors}
    \label{fig:hardware_setup}
\end{figure}

For the real-world experiments, we used the Agilex Hunter 2.0 platform\footnote[2]{\url{https://robosavvy.co.uk/agilex-hunter-2b.html} \label{hunter}}, a versatile and durable robotic platform optimised for outdoor environments. The robot was outfitted with several critical hardware components, including a Trimble RTK-GPS\footnote[3]{\url{https://www.trimble.com/en/solutions/technologies/positioning} \label{gps}} for high-precision localization, essential for maintaining accurate navigation in agricultural settings. An Intel NUC i7-10710U\footnote[4]{\url{https://www.intel.com/content/www/us/en/support/} \label{pc}} served as the onboard computer, managing the complex processing tasks associated with navigation, sensor integration, and control. Fig. \ref{fig:hardware_setup} illustrates the complete hardware configuration.

The navigation stack was implemented in C++ within the ROS2 framework, adhering to ROS2’s plugin-based architecture. Each navigation plugin was designed to align with ROS2's standards, allowing seamless integration with other ROS2 components and supporting future updates or adaptations. This modular approach enabled us to develop a flexible and efficient navigation solution suitable for both simulation and real-world testing, facilitating smooth transitions between the two environments.

\subsection{Performance evaluation of the proposed RTEB planner}

\begin{figure}[ht!]
    \centering
    % \vspace{0.3cm}
    \includegraphics[width=1.0\linewidth]{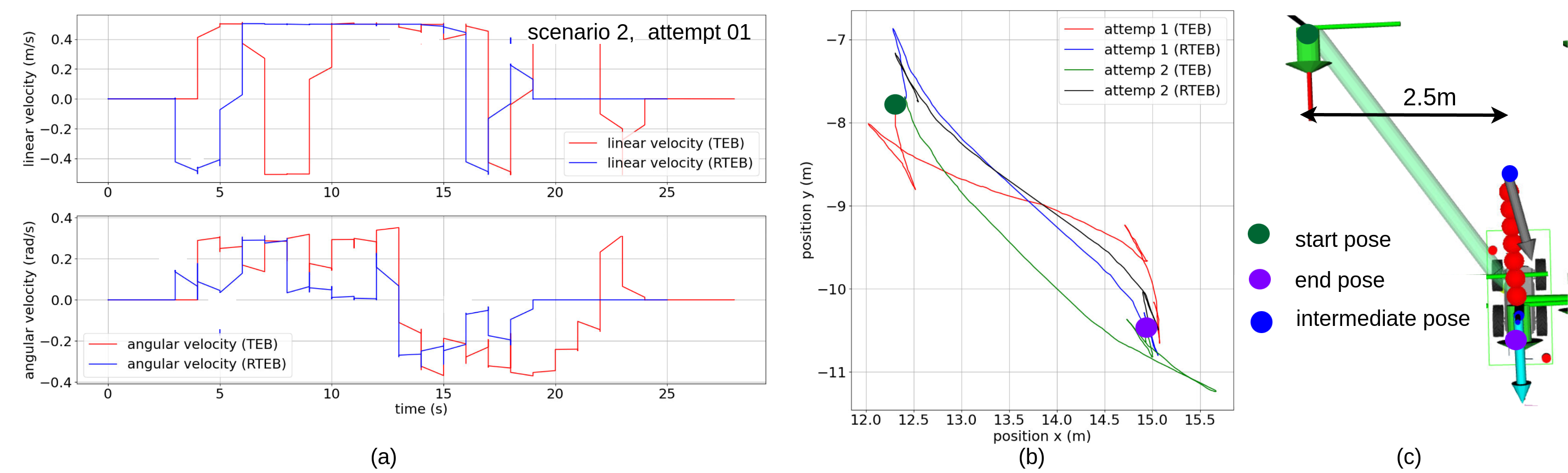}
    \caption{The comparison of planning performance evaluates the goal alignment estimation of RTEB in relation to TEB. In this scenario (the second scenario in Table\ref{table:result_teb_goal_align}, which involves a 2.5m displacement between the starting and target poses), the starting and target poses remain constant. The robot is planned and navigated twice using each method. RTEB generates more consistent trajectories, while TEB exhibits variability over time.}
    \label{fig:goal_alignment_test}
\end{figure}

In the first experiment, we assessed the goal alignment capabilities of the RTEB planner in comparison with TEB across several defined scenarios. This experiment involved a controlled setup where the start and target orientations were maintained constant, while the displacement between them varied, simulating a parallel parking task (as depicted in Fig.~\ref{fig:goal_alignment_test}). Displacement values ranged from 2 m to 4 m, increasing in 0.5 m steps, with each scenario corresponding to specific displacement values (e.g., scenario 1 is set to 2 m, and scenario 5 to 4 m), as shown in Table \ref{table:result_teb_goal_align}. We established a tolerance threshold of 0.2 m for XY positioning and 0.1 rad for yaw to ensure meaningful comparisons of goal alignment accuracy. To evaluate performance, we measured three main criteria: the traverse distance, the traversal time ($T$), and control effort \(\left( \int_{0}^{T} \|\mathbf{u}(t)\|^2 \, dt \right)\), where the control input vector \(\mathbf{u}\) comprises linear velocity $v$ and angular velocity $\omega$ values over time. Each scenario was repeated ten times to account for variability, and we present the averaged results with standard deviations in Table \ref{table:result_teb_goal_align}.

\begin{table}[h!t]
\centering
\caption{Comparison of goal alignment performance improvement in RTEB compared to TEB (values are given as $\mu \pm \sigma$)}
\label{table:result_teb_goal_align}
\begin{tblr}{
  row{1} = {c},
  row{3} = {c},
  row{4} = {c},
  row{7} = {c},
  cell{1}{2} = {c=2}{},
  cell{1}{4} = {c=2}{},
  cell{1}{6} = {c=2}{},
  cell{3}{3} = {Silver},
  cell{3}{5} = {Silver},
  cell{3}{7} = {Silver},
  cell{4}{3} = {Silver},
  cell{4}{5} = {Silver},
  cell{4}{7} = {Silver},
  cell{5}{3} = {Silver},
  cell{5}{5} = {Silver},
  cell{5}{7} = {Silver},
  cell{6}{3} = {Silver},
  cell{6}{5} = {Silver},
  cell{6}{7} = {Silver},
  cell{7}{3} = {Silver},
  cell{7}{5} = {Silver},
  cell{7}{7} = {Silver},
  vline{2-3,5} = {1}{},
  vline{2,4,6} = {2}{},
  vline{2-8} = {3-7}{},
  hline{1,3} = {-}{},
  hline{2} = {2-7}{},
}
Scenario & {Traverse Distance\\~(m)~~} &             & {Traverse Time\\~(s)~~} &              & Control Effort~~ &              \\
         & TEB                         & ~ ~RTEB     & ~ TEB                   & ~RTEB        & ~ TEB            & ~RTEB        \\
1        & 3.00 ± 1.15                 & 2.86 ± 0.87 & 21.52 ± 3.95            & 18.00 ± 2.88 & 4.78 ± 1.25      & 3.82 ± 1.21  \\
2        & 4.78 ± 1.32                 & 4.51 ± 1.25 & 33.00 ± 1.25            & 26.00 ± 2.12 & 10.36 ± 1.55     & 6.41 ± 1.38  \\
~ ~ ~3   & 5.41 ± 1.36            & 5.95 ± 1.28 & 37.54 ± 2.10            & 32.00 ± 3.00 & 18.12 ± 2.48     & 13.00 ± 1.32 \\
~ ~ ~4   & 8.54 ± 1.30                  & 6.55 ± 0.74 & 49.61 ± 4.20            & 41.00 ± 1.05 & 24.77 ± 1.65     & 18.79 ± 1.50 \\
5        & 11.17 ± 1.41                & 8.97 ± 0.81 & 55.21 ± 1.50            & 48.00 ± 2.45 & 27.58 ± 2.72     & 8.96 ± 1.60  
\end{tblr}
\end{table}

The results in Table \ref{table:result_teb_goal_align} demonstrate a significant improvement in RTEB's performance over TEB across three key metrics: traverse distance, traverse time, and control effort. Paired t-tests~\cite{hsu2014paired} were used to assess the statistical significance of the differences between RTEB and TEB for each metric. For the traverse time, the p-value was \( p = 0.002 \), indicating a statistically significant difference, suggesting that RTEB consistently reaches the target faster than TEB. This time efficiency could be advantageous in applications where rapid task completion is essential. Similarly, the control effort metric showed a significant difference, with a p-value of \( p = 0.005 \). Control effort, calculated as the integral of squared control inputs, reflects the smoothness and energy expenditure of the robot’s movement. This result implies that RTEB and TEB differ in terms of motion smoothness and energy efficiency, with RTEB offering a more refined control strategy. In contrast, for traverse distance, the p-value was \( p = 0.209 \), indicating no significant difference in the total distance travelled. This suggests that RTEB and TEB are equally effective in minimizing the path length required to reach the goal.

Overall, while both algorithms perform similarly in traverse distance, RTEB and TEB display significant differences in traverse time and control effort, which could impact their suitability for specific applications. Additionally, RTEB consistently demonstrates lower mean values and standard deviations across all metrics, indicating improved efficiency and reduced variability in goal alignment scenarios. For example, in Scenario 1, RTEB reduced the traverse time by an average of 16.4\%, alongside a notable decrease in control effort. This trend is consistent across various displacement scenarios, with RTEB achieving shorter distances, times, and more efficient control inputs, particularly in Scenarios 4 and 5. As shown in Fig. \ref{fig:goal_alignment_test}, RTEB provides superior goal alignment, resulting in more consistent and reliable trajectories. These findings support RTEB as a robust approach for applications demanding precise and efficient goal alignment.

\begin{figure}[ht!]
    \centering
    \includegraphics[width=1.0\linewidth]{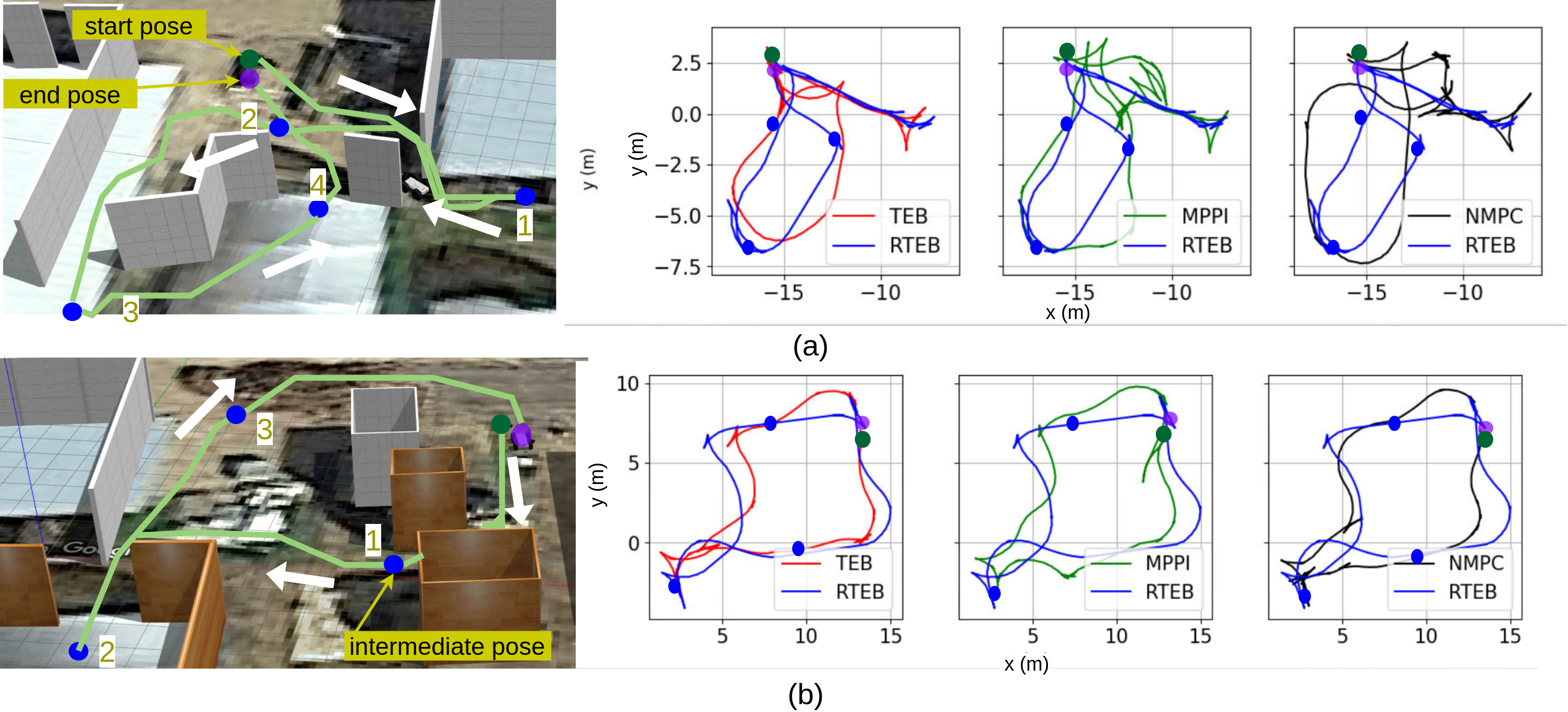}
    \caption{Performance evaluation of RTEB, TEB, NMPC, and MPPI in randomised dense obstacle simulated environments}
    \label{fig:checking_gaps}
\end{figure}

In this second experiment, we evaluated the performance of RTEB, TEB, NMPC, and MPPI  in randomised dense obstacle scenarios across five simulated environments (see Fig. \ref{fig:checking_gaps} for examples). Each environment featured four distinct start and goal position pairs, with twenty trials conducted per pair. These environments comprised static obstacles randomly arranged in a confined 3D space, creating narrow gaps and challenging navigation conditions. The goal was to assess each planner's efficiency in traversing narrow gaps while minimizing collisions and path deviations. To quantify environmental complexity, we used the average width of the two narrowest gaps as a benchmark. This metric was computed by measuring gap widths between adjacent obstacles, sorting them in ascending order, and averaging the two smallest values. The resulting narrow gap widths ranged from 1.5 m to 2.5 m, highlighting the most difficult passages for navigation. By focusing on these critical gaps, the complexity metric ensures a representative assessment of each algorithm's performance under challenging conditions. The results of this experiment are summarised in Table~\ref{tab:narrow_gap}.

\begin{table}[ht!]
\centering
\caption{Performance comparison of trajectory planning methods in scenarios involving narrow gaps within the environment}
\label{tab:narrow_gap}
\begin{tblr}{
  column{even} = {c},
  column{3} = {c},
  column{5} = {c},
  cell{2}{2} = {Silver},
  cell{2}{3} = {fg=MineShaft},
  cell{2}{4} = {fg=MineShaft},
  cell{3}{2} = {Silver},
  cell{4}{5} = {Silver},
  cell{5}{2} = {Silver},
  cell{6}{2} = {Silver},
  cell{7}{2} = {Silver},
  vline{2-5} = {-}{},
  hline{1-7} = {-}{},
}
\textbf{Metric}                    & \textbf{RTEB~}            & \textbf{TEB~}    & \textbf{NMPC~} & \textbf{MPPI~} \\
Success Rate (\%)                  & 90                        & 80               & 75             & 70             \\
{Path Efficiency\\~(Path Ratio)}   & 0.91                      & 0.72             & 0.68           & 0.62           \\
{Maximum Planning\\Frequency (Hz)} & 17                        & 13               & 12             & 20             \\
Handling Narrow Gaps               & Good                      & Moderate         & Moderate       & Moderate       \\
Orientation Flexibility            & Bidirectional (Hybrid A*) & Forward-Oriented & Bidirectional  & Bidirectional  \\
{Adaptability to \\Complexity}     & High                      & Medium           & Medium         & Medium         
\end{tblr}
\end{table}         

Across several key metrics, RTEB consistently outperformed the other methods, demonstrating its robustness and efficiency in complex navigation scenarios. In terms of success rate, the percentage of trials in which the planner completed the task without collisions or failures, RTEB achieved 90\%, significantly higher than TEB (80\%), NMPC (75\%), and MPPI (70\%). This indicates that RTEB is more reliable in successfully navigating dense obstacle environments with narrow gaps. Such a high success rate underscores its ability to effectively plan and execute feasible trajectories even under challenging conditions. Path efficiency, the ratio of the actual path length taken by the planner to the shortest possible path length, further reinforces RTEB’s superiority. It achieved a ratio of 0.91, closely approximating the feasible path, while TEB (0.72), NMPC (0.68), and MPPI (0.62) displayed lower efficiency. This demonstrates that RTEB generates shorter, smoother trajectories, minimizing deviations from the ideal path while navigating narrow gaps.

The maximum planning frequency, the highest rate at which the planner can generate updated trajectories, reflects the computational performance of each method. RTEB operated at 17 Hz, outperforming TEB (13 Hz) and NMPC (12 Hz), though MPPI achieved the highest frequency (20 Hz). However, while MPPI’s higher frequency may enable rapid re-planning, its lower path efficiency and success rate suggest it struggles with the complexities of narrow gaps, making RTEB a more balanced choice for such environments. In terms of handling narrow gaps, a qualitative assessment of the planner's ability to navigate through tight spaces or obstacles with minimal clearance: "Good," and "Moderate" indicates the relative effectiveness in such scenarios, RTEB is rated "Good," while the other planners are rated "Moderate." Orientation flexibility highlights RTEB and MPPI as bidirectional planners, with RTEB leveraging Hybrid A* for enhanced flexibility, whereas TEB is constrained to forward-oriented planning. Finally, adaptability to complexity, which evaluates how well the planner performs in environments with high levels of obstacle density, shows RTEB rated "High," while the others are "Medium," underscoring RTEB's superior robustness when increasing obstacle density.

\begin{figure}[ht!]
    \centering
    \includegraphics[width=1.0\linewidth]{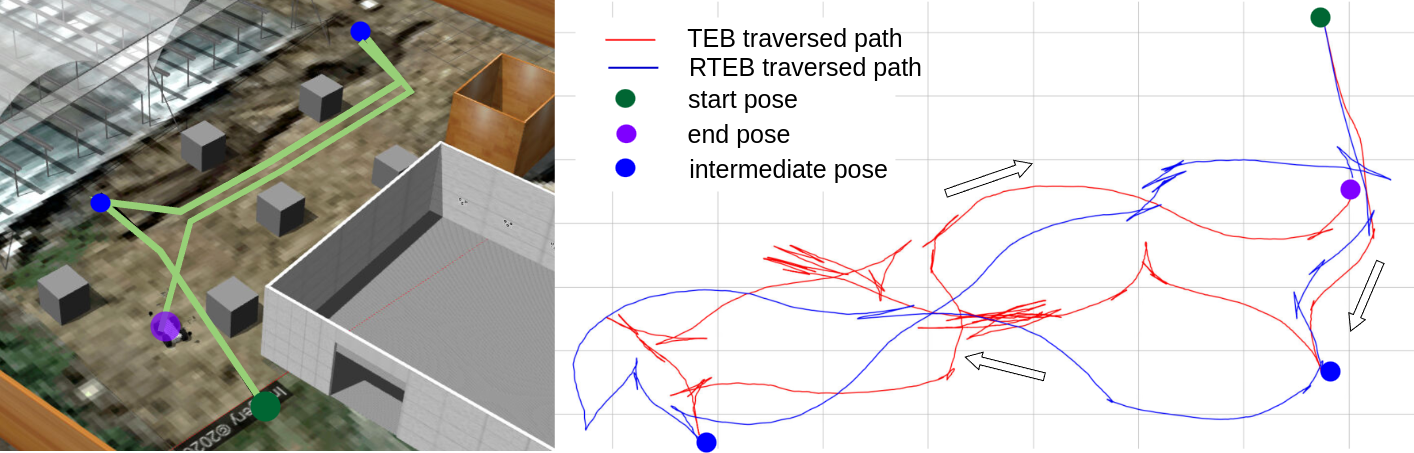}
    \caption{Left: example of the simulated experiment setup in Gazebo, showcasing the environment with varying obstacle densities used to validate TEB versus RTEB performance. Right: The final trajectories for TEB and RTEB methods illustrate the paths navigated through the obstacle-laden environment. }
    \label{fig:try_sim_test}
\end{figure}

The goal of the final experiment is to comprehensively compare the trajectory planning performance of three different planners: TEB planner, RTEB (the proposed planner), and NMPC~\cite{rosmann2021online}. This comparison was conducted in both simulated and real environments with obstacle densities of 20\%, 30\%, 40\%, and 50\% (Fig.\ref{fig:try_sim_test}), complemented by maintaining a minimum distance of 2.5 m between obstacles. Obstacle density was estimated as the ratio of free space before placing obstacles to the free space after placing new obstacles, expressed as a percentage. A common reference path is used for all test cases to ensure comparability of results across different obstacle densities and environments. The Gazebo provides a controlled setting in the simulated environment where obstacle densities are systematically varied. The same environmental configurations and obstacle densities are replicated in the real-world setup to validate the simulation results. Performance is assessed based on traverse distance, traverse time ($T$), control effort ($ \int_{0}^{T} |\mathbf{u}(t)|^2 \, dt$), speed range (minimum and maximum speeds), and maximum planning frequency (the average time taken per planning iteration). Each experiment was conducted five times, and the average values are presented in Table~\ref{table:result_teb_rteb}. 

\begin{table}[ht!]
\centering
\caption{Comparison of trajectory planning performance across various environmental settings in real and simulated setups}
\label{table:result_teb_rteb}
\begin{tabular}{l|c|c|c|c|c|c|c|c|c|c} 
\hline
\multicolumn{2}{l|}{~ ~ Attempt}                                                                                        & \multicolumn{3}{c|}{\begin{tabular}[c]{@{}c@{}}Traverse Distance\\~{[}m]~~\end{tabular}}                                            & \multicolumn{3}{c|}{\begin{tabular}[c]{@{}c@{}}Traverse Time\\~{[}s]~~\end{tabular}}                                & \multicolumn{3}{c}{Control Effort~~}                                                                               \\ 
\hline
                                      & \multicolumn{1}{l|}{\begin{tabular}[c]{@{}l@{}}obstacle~\\density\end{tabular}} & \multicolumn{1}{l}{TEB}                    & \multicolumn{1}{l}{~RTEB}                  & \multicolumn{1}{l|}{NMPC}                 & \multicolumn{1}{l}{TEB}                    & \multicolumn{1}{l}{RTEB}                   & \multicolumn{1}{l|}{NMPC} & \multicolumn{1}{l}{TEB}                   & \multicolumn{1}{l}{RTEB}                   & \multicolumn{1}{l}{NMPC}  \\ 
\hhline{~----------}
\multirow{4}{*}{\rotatebox{90}{Simulation}} & 20\%                                                                            & {\cellcolor[rgb]{0.753,0.753,0.753}}65.32  & {\cellcolor[rgb]{0.753,0.753,0.753}}65.33  & 71.82                                     & {\cellcolor[rgb]{0.753,0.753,0.753}}167.00 & 171.00                                     & 191.00                    & {\cellcolor[rgb]{0.753,0.753,0.753}}70.70 & {\cellcolor[rgb]{0.753,0.753,0.753}}69.20  & 78.31                     \\
                                      & 30\%                                                                            & 76.59                                      & {\cellcolor[rgb]{0.753,0.753,0.753}}69.21  & 79.29                                     & 230.00                                     & {\cellcolor[rgb]{0.753,0.753,0.753}}178.00 & 218.00                    & 83.39                                     & {\cellcolor[rgb]{0.753,0.753,0.753}}77.45  & 93.21                     \\
                                      & 40\%                                                                            & 115.54                                     & {\cellcolor[rgb]{0.753,0.753,0.753}}77.11  & 97.21                                     & 352.00                                     & {\cellcolor[rgb]{0.753,0.753,0.753}}216.00 & 249.00                    & 140.77                                    & {\cellcolor[rgb]{0.753,0.753,0.753}}79.13  & 168.34                    \\
                                      & ~50\%                                                                           & 145.54                                     & {\cellcolor[rgb]{0.753,0.753,0.753}}120.11 & ~ NaN                                     & 432.02                                     & {\cellcolor[rgb]{0.753,0.753,0.753}}257.00 & ~ NaN                     & 196.40                                    & {\cellcolor[rgb]{0.753,0.753,0.753}}100.11 & ~ NaN                     \\ 
\hline
\multirow{4}{*}{\rotatebox{90}{Real-world}} & 20\%                                                                            & {\cellcolor[rgb]{0.753,0.753,0.753}}65.89  & 66.40                                      & 73.12                                     & {\cellcolor[rgb]{0.753,0.753,0.753}}168.10 & 170.00                                     & 196.00                    & {\cellcolor[rgb]{0.753,0.753,0.753}}67.82 & 68.87                                      & 70.78                     \\
                                      & 30\%                                                                            & 78.9                                       & {\cellcolor[rgb]{0.753,0.753,0.753}}70.39  & 81.29                                     & 239.00                                     & {\cellcolor[rgb]{0.753,0.753,0.753}}181.00 & 226.00                    & 85.21                                     & {\cellcolor[rgb]{0.753,0.753,0.753}}78.29  & 95.69                     \\
                                      & 40\%                                                                            & 112.16                                     & {\cellcolor[rgb]{0.753,0.753,0.753}}78.28  & 100.64                                    & 347.00                                     & {\cellcolor[rgb]{0.753,0.753,0.753}}220.00 & 256.00                    & 135.28                                    & {\cellcolor[rgb]{0.753,0.753,0.753}}81.37  & 173.25                    \\
                                      & 50\%                                                                            & 162.54                                     & {\cellcolor[rgb]{0.753,0.753,0.753}}138.11 & ~ NaN                                     & ~448.02                                    & {\cellcolor[rgb]{0.753,0.753,0.753}}269.00 & ~ NaN                     & 212.40                                    & {\cellcolor[rgb]{0.753,0.753,0.753}}114.11 & ~ NaN                     \\ 
\hline
\multicolumn{2}{l|}{~ ~ Attempt}                                                                                        & \multicolumn{3}{c|}{\begin{tabular}[c]{@{}c@{}}Speed (min, max)~~\\{[}m/s]~\end{tabular}}                                           & \multicolumn{3}{c}{\begin{tabular}[c]{@{}c@{}}MaximumPlanningFrequency \\{[}Hz]~\end{tabular}}                      & \multicolumn{3}{c}{}                                                                                               \\ 
\cline{1-8}
                                      & \multicolumn{1}{l|}{\begin{tabular}[c]{@{}l@{}}obstacle~\\density\end{tabular}} & \multicolumn{1}{l}{TEB}                    & \multicolumn{1}{l}{RTEB}                   & \multicolumn{1}{l|}{NMPC}                 & \multicolumn{1}{l}{TEB}                    & \multicolumn{1}{l}{RTEB}                   & \multicolumn{1}{l}{NMPC}  & \multicolumn{3}{l}{}                                                                                               \\ 
\hhline{~-------~~~}
\multirow{4}{*}{\rotatebox{90}{Simulation}} & 20\%                                                                            & {\cellcolor[rgb]{0.753,0.753,0.753}}±0.51  & ±0.53                                      & 0.52                                      & {\cellcolor[rgb]{0.753,0.753,0.753}}20±3   & {\cellcolor[rgb]{0.753,0.753,0.753}}20±2   & \multicolumn{1}{c}{17±5}  & \multicolumn{1}{c}{}                      & \multicolumn{1}{c}{}                       &                           \\
                                      & 30\%                                                                            & {\cellcolor[rgb]{0.753,0.753,0.753}}±0.81  & ±0.76                                      & ±0.76                                     & {\cellcolor[rgb]{0.753,0.753,0.753}}19±3   & 17±5                                       & \multicolumn{1}{c}{12±5}  & \multicolumn{1}{c}{}                      & \multicolumn{1}{c}{}                       &                           \\
                                      & 40\%                                                                            & {\cellcolor[rgb]{0.753,0.753,0.753}}±0.54  & ±0.54                                      & ±0.78                                     & {\cellcolor[rgb]{0.753,0.753,0.753}}17±4   & 16±4                                       & \multicolumn{1}{c}{~9±5}  & \multicolumn{1}{c}{}                      & \multicolumn{1}{c}{}                       &                           \\
                                      & ~50\%                                                                           & {\cellcolor[rgb]{0.753,0.753,0.753}}±0.52  & NaN                                        & ~ NaN                                     & {\cellcolor[rgb]{0.753,0.753,0.753}}15±3   & 14±5                                       & \multicolumn{1}{c}{~ NaN} & \multicolumn{1}{c}{}                      & \multicolumn{1}{c}{}                       & ~~                        \\ 
\hhline{--------~~~}
\multirow{4}{*}{\rotatebox{90}{Real-world}} & 20\%                                                                            & ±0.55                                      & ±0.55                                      & {\cellcolor[rgb]{0.753,0.753,0.753}}±0.54 & {\cellcolor[rgb]{0.753,0.753,0.753}}19±3   & 19±5                                       & \multicolumn{1}{c}{15±5}  & \multicolumn{1}{c}{}                      & \multicolumn{1}{c}{}                       &                           \\
                                      & 30\%                                                                            & {\cellcolor[rgb]{0.753,0.753,0.753}}±0.55  & {\cellcolor[rgb]{0.753,0.753,0.753}}±0.55  & {\cellcolor[rgb]{0.753,0.753,0.753}}±0.55 & {\cellcolor[rgb]{0.753,0.753,0.753}}17±3   & 14±5                                       & \multicolumn{1}{c}{11±5}  & \multicolumn{1}{c}{}                      & \multicolumn{1}{c}{}                       &                           \\
                                      & 40\%                                                                            & {\cellcolor[rgb]{0.753,0.753,0.753}}±0.53  & ±0.54                                      & ±0.58                                     & {\cellcolor[rgb]{0.753,0.753,0.753}}15±4   & 11±5                                       & \multicolumn{1}{c}{7±5}   & \multicolumn{1}{c}{}                      & \multicolumn{1}{c}{}                       &                           \\
                                      & 50\%                                                                            & {\cellcolor[rgb]{0.753,0.753,0.753}}~±0.83 & ±0.52                                      & ~ NaN                                     & {\cellcolor[rgb]{0.753,0.753,0.753}}~15±3  & 14±5                                       & \multicolumn{1}{c}{~ NaN} & \multicolumn{1}{c}{}                      & \multicolumn{1}{c}{}                       & ~~                        \\
\hhline{--------~~~}
\end{tabular}
\begin{tablenotes}
   \item[*] NaN: the planner is unable to reach the target location and continues to loop around.
\end{tablenotes}
\end{table}

Table \ref{table:result_teb_rteb} reveals that RTEB consistently outperforms TEB, particularly in environments with increasing obstacle density, where RTEB’s stability is reflected by its lower standard deviation values across traverse distance, time, and control effort metrics. This stability is crucial for applications in dynamic and densely populated environments, as it implies that RTEB provides more predictable and reliable trajectory outcomes under challenging conditions. Additionally, RTEB’s ability to maintain a comparable or slightly higher planning frequency than TEB across most scenarios indicates its computational efficiency despite handling complex path adjustments.
For example, all planners perform similarly at a low obstacle density of 20\%, with TEB and RTEB achieving nearly identical distances of around 63 m, while NMPC covers a slightly longer distance of 71.82 m. As the obstacle density increases, NMPC tends to cover longer distances, particularly at 40\% density, where it records 97.21 m compared to RTEB’s 77.11 m, which is the shortest distance among the planners. In real environments, the planners show similar trends, with NMPC consistently covering the longest distance, particularly at 40\% density (100.64 m), suggesting a potential trade-off between distance and obstacle avoidance. Therefore, NMPC, while occasionally producing smoother trajectories (noted by tighter control effort and speed variation), experiences a decline in planning frequency and robustness, especially at higher obstacle densities. This decrease suggests that NMPC, although useful in smoother, less complex setups, may not be as suited for dynamic environments requiring rapid, adaptive path adjustments.

For a traverse time, the analysis reveals that the time required increases significantly with obstacle density across all planners. However, RTEB generally results in shorter traverse times compared to TEB and NMPC, particularly in denser environments. For instance, at 40\% obstacle density in simulated environments, RTEB and TEB take the average of 216 s and 352 s. This pattern holds in real environments, where RTEB consistently shows shorter times, with a notable gap at higher densities, such as 220 s for RTEB versus 347 s for TEB at 40\% density. When considering control effort, which reflects the intensity of manoeuvres required to avoid obstacles, it is observed that control effort increases with obstacle density across all planners. NMPC generally demands the highest control effort, especially at high densities, reflecting its more aggressive obstacle avoidance strategy. In simulated environments, NMPC’s control effort reaches 168.34 at 40\% density, while RTEB shows relatively stable and lower control efforts. Similar trends are observed in real environments, with NMPC again requiring the highest control effort (173.25 at 40\% density), indicating its resource-intensive nature compared to TEB and RTEB, which demonstrate more efficiency. 

This suggests that the modifications in RTEB enhance its ability to handle dense environments more efficiently. Overall, RTEB’s balance of efficiency and adaptability makes it a strong candidate for real-world applications where reliability and computational load are both critical factors, particularly in more complex environments with higher obstacle densities. In obstacle-free environments, there is no notable difference between TEB and RTEB, as the recovery mechanism is unnecessary. As a result, within the RTEB planner, the recovery mechanism remains inactive, and planning proceeds without it, which is advantageous in terms of computational cost.   

\section{Conclusion}

This paper introduces RTEB, an enhanced version of the Timed Elastic Band (TEB) planner, designed to outperform both TEB and Nonlinear Model Predictive Control (NMPC) in terms of traverse time and control effort, particularly in complex and dense environments. One of the key innovations of RTEB is the incorporation of a resilient trajectory generation method, which leverages a hybrid A* algorithm to reinitialise the TEB planner when it encounters failure situations. This capability significantly enhances the planner’s consistency and speed, particularly in environments with high clutter and dynamic obstacles, where traditional planning methods often struggle to maintain efficient and feasible solutions.

Additionally, RTEB integrates a soft constraints-based smoothing utility that further refines the generated trajectories. This utility ensures that the resulting paths are not only efficient but also safe, smooth, and feasible. The soft constraints allow for the inclusion of dynamic and environmental factors that affect the robot's motion, such as changes in velocity or unforeseen obstacle movements. This refinement ensures that the robot can navigate through complex, cluttered spaces with minimal control effort, while also adhering to safety constraints. These features collectively contribute to the robust and flexible nature of RTEB, making it an ideal choice for autonomous navigation in unknown, dynamically changing environments.

For the validation of the proposed RTEB planner, we utilised the ROS2 navigation stack, exploiting the available local planners, specifically TEB and NMPC. While several recent trajectory planning approaches, such as the Spatial-Temporal Trajectory Planner \cite{han2023efficient}, OBTPAP \cite{li2021optimization}, and DL-IAPS+PJSO \cite{zhou2020autonomous}, have demonstrated strong performance in ROS1 environments, their adaptation to ROS2 has proven to be time-consuming due to various system constraints. Consequently, we opted to validate our RTEB approach against the more widely adopted TEB and NMPC planners. In future work, we aim to validate RTEB against these recent methods once their ROS2 implementations are available, expanding the comparison to understand better RTEB's performance in relation to these state-of-the-art planners.

Furthermore, the resilient trajectory generation approach introduced in this paper is highly adaptable and can be integrated with any local planning method, offering flexibility for various robot configurations and environments. Looking ahead, we plan to integrate RTEB with Nonlinear Model Predictive Control (NMPC), further enhancing its capabilities. However, the integration of RTEB with NMPC presents certain challenges. Unlike TEB, which utilises a variable receding horizon for planning, NMPC relies on a fixed receding horizon, which may affect the implementation of the resilient planning strategy. To address this, we foresee the need for a cascade planning approach that harmonises soft and hard constraints within the NMPC framework, enabling real-time performance without compromising trajectory quality. This research direction will be a focus of future work, aiming to bridge the gap between these two powerful planning methods and enhance the robustness and efficiency of autonomous navigation systems.

\section*{Acknowledgements}
This work was supported by the Innovate UK-funded project Agri-OpenCore [grant number 10041179].

\bibliographystyle{apalike}
\bibliography{sample}

\begin{thebibliography}{}

\bibitem[Allozi et~al., 2022]{allozi2022feasibility}
Allozi, E., Yilmaz, A., Ervan, O., and Temeltas, H. (2022).
\newblock Feasibility analysis of path planning algorithms.
\newblock In {\em 2022 International Conference on INnovations in Intelligent SysTems and Applications (INISTA)}, pages 1--6. IEEE.

\bibitem[Astudillo et~al., 2024]{astudillo2024rapid}
Astudillo, A., Florez, A., Decr{\'e}, W., and Swevers, J. (2024).
\newblock Rapid deployment of model predictive control for robotic systems: From impact to ros 2 through code generation.
\newblock In {\em 2024 IEEE 18th International Conference on Advanced Motion Control (AMC)}, pages 1--6. IEEE.

\bibitem[Chaplot et~al., 2020]{chaplot2020semantic}
Chaplot, D.~S., Jiang, H., Gupta, S., and Gupta, A. (2020).
\newblock Semantic curiosity for active visual learning.
\newblock In {\em Computer Vision--ECCV 2020: 16th European Conference, Glasgow, UK, August 23--28, 2020, Proceedings, Part VI 16}, pages 309--326. Springer.

\bibitem[Chen et~al., 2019]{chen2019learning}
Chen, T., Gupta, S., and Gupta, A. (2019).
\newblock Learning exploration policies for navigation.
\newblock {\em arXiv preprint arXiv:1903.01959}.

\bibitem[Chung et~al., 2022]{chung2022distributed}
Chung, Y.~M., Youssef, H., and Roidl, M. (2022).
\newblock Distributed timed elastic band (dteb) planner: Trajectory sharing and collision prediction for multi-robot systems.
\newblock In {\em 2022 International Conference on Robotics and Automation (ICRA)}, pages 10702--10708. IEEE.

\bibitem[Delsart and Fraichard, 2008]{delsart2008reactive}
Delsart, V. and Fraichard, T. (2008).
\newblock Reactive trajectory deformation to navigate dynamic environments.
\newblock In {\em European Robotics Symposium 2008}, pages 233--241. Springer.

\bibitem[Dolgov et~al., 2008]{dolgov2008practical}
Dolgov, D., Thrun, S., Montemerlo, M., and Diebel, J. (2008).
\newblock Practical search techniques in path planning for autonomous driving.
\newblock {\em Ann Arbor}, 1001(48105):18--80.

\bibitem[Fang et~al., 2023]{fang2023generalization}
Fang, K., Yin, P., Nair, A., Walke, H.~R., Yan, G., and Levine, S. (2023).
\newblock Generalization with lossy affordances: Leveraging broad offline data for learning visuomotor tasks.
\newblock In {\em Conference on Robot Learning}, pages 106--117. PMLR.

\bibitem[Faust et~al., 2018]{faust2018prm}
Faust, A., Oslund, K., Ramirez, O., Francis, A., Tapia, L., Fiser, M., and Davidson, J. (2018).
\newblock Prm-rl: Long-range robotic navigation tasks by combining reinforcement learning and sampling-based planning.
\newblock In {\em 2018 IEEE international conference on robotics and automation (ICRA)}, pages 5113--5120. IEEE.

\bibitem[Fox et~al., 1997]{fox1997dynamic}
Fox, D., Burgard, W., and Thrun, S. (1997).
\newblock The dynamic window approach to collision avoidance.
\newblock {\em IEEE Robotics \& Automation Magazine}, 4(1):23--33.

\bibitem[Ge et~al., 2023]{ge2023efficient}
Ge, L., Zhao, Y., Zhong, S., Shan, Z., and Guo, K. (2023).
\newblock Efficient nonlinear model predictive motion controller for autonomous vehicles from standstill to extreme conditions based on split integration method.
\newblock {\em Control Engineering Practice}, 141:105720.

\bibitem[Gervet et~al., 2023]{gervet2023navigating}
Gervet, T., Chintala, S., Batra, D., Malik, J., and Chaplot, D.~S. (2023).
\newblock Navigating to objects in the real world.
\newblock {\em Science Robotics}, 8(79):eadf6991.

\bibitem[Ghzouli et~al., 2023]{ghzouli2023behavior}
Ghzouli, R., Berger, T., Johnsen, E.~B., Wasowski, A., and Dragule, S. (2023).
\newblock Behavior trees and state machines in robotics applications.
\newblock {\em IEEE Transactions on Software Engineering}, 49(9):4243--4267.

\bibitem[Han et~al., 2023]{han2023efficient}
Han, Z., Wu, Y., Li, T., Zhang, L., Pei, L., Xu, L., Li, C., Ma, C., Xu, C., Shen, S., et~al. (2023).
\newblock An efficient spatial-temporal trajectory planner for autonomous vehicles in unstructured environments.
\newblock {\em IEEE Transactions on Intelligent Transportation Systems}.

\bibitem[Horne and Cole~Smith, 2005]{horne2005dynamic}
Horne, J.~A. and Cole~Smith, J. (2005).
\newblock Dynamic programming algorithms for the conditional covering problem on path and extended star graphs.
\newblock {\em Networks: An International Journal}, 46(4):177--185.

\bibitem[Houshyari and Sezer, 2022]{houshyari2022new}
Houshyari, H. and Sezer, V. (2022).
\newblock A new gap-based obstacle avoidance approach: follow the obstacle circle method.
\newblock {\em Robotica}, 40(7):2231--2254.

\bibitem[Hsu and Lachenbruch, 2014]{hsu2014paired}
Hsu, H. and Lachenbruch, P.~A. (2014).
\newblock Paired t test.
\newblock {\em Wiley StatsRef: statistics reference online}.

\bibitem[Huajian et~al., 2024]{huajian2024sample}
Huajian, L., Wei, D., Shouren, M., Chao, W., and Yongzhuo, G. (2024).
\newblock Sample-efficient learning-based dynamic environment navigation with transferring experience from optimization-based planner.
\newblock {\em IEEE Robotics and Automation Letters}.

\bibitem[Kadian et~al., 2020]{kadian2020sim2real}
Kadian, A., Truong, J., Gokaslan, A., Clegg, A., Wijmans, E., Lee, S., Savva, M., Chernova, S., and Batra, D. (2020).
\newblock Sim2real predictivity: Does evaluation in simulation predict real-world performance?
\newblock {\em IEEE Robotics and Automation Letters}, 5(4):6670--6677.

\bibitem[Kazim et~al., 2024]{kazim2024recent}
Kazim, M., Hong, J., Kim, M.-G., and Kim, K.-K.~K. (2024).
\newblock Recent advances in path integral control for trajectory optimization: An overview in theoretical and algorithmic perspectives.
\newblock {\em Annual Reviews in Control}, 57:100931.

\bibitem[Khazatsky et~al., 2021]{khazatsky2021can}
Khazatsky, A., Nair, A., Jing, D., and Levine, S. (2021).
\newblock What can i do here? learning new skills by imagining visual affordances.
\newblock In {\em 2021 IEEE International Conference on Robotics and Automation (ICRA)}, pages 14291--14297. IEEE.

\bibitem[Kim et~al., 2022]{kim2022smooth}
Kim, T., Park, G., Kwak, K., Bae, J., and Lee, W. (2022).
\newblock Smooth model predictive path integral control without smoothing.
\newblock {\em IEEE Robotics and Automation Letters}, 7(4):10406--10413.

\bibitem[Kr{\"a}mer et~al., 2020]{kramer2020model}
Kr{\"a}mer, M., R{\"o}smann, C., Hoffmann, F., and Bertram, T. (2020).
\newblock Model predictive control of a collaborative manipulator considering dynamic obstacles.
\newblock {\em Optimal Control Applications and Methods}, 41(4):1211--1232.

\bibitem[Kulathunga et~al., 2022a]{kulathunga2022trajectory}
Kulathunga, G., Devitt, D., and Klimchik, A. (2022a).
\newblock Trajectory tracking for quadrotors: An optimization-based planning followed by controlling approach.
\newblock {\em Journal of Field Robotics}, 39(7):1001--1011.

\bibitem[Kulathunga et~al., 2022b]{kulathunga2022optimization}
Kulathunga, G., Hamed, H., Devitt, D., and Klimchik, A. (2022b).
\newblock Optimization-based trajectory tracking approach for multi-rotor aerial vehicles in unknown environments.
\newblock {\em IEEE Robotics and Automation Letters}, 7(2):4598--4605.

\bibitem[Kulathunga and Klimchik, 2023]{kulathunga2023survey}
Kulathunga, G. and Klimchik, A. (2023).
\newblock Survey on motion planning for multirotor aerial vehicles in plan-based control paradigm.
\newblock {\em Remote Sensing}, 15(21):5237.

\bibitem[K{\"u}mmerle et~al., 2011]{kummerle2011g}
K{\"u}mmerle, R., Grisetti, G., Strasdat, H., Konolige, K., and Burgard, W. (2011).
\newblock g 2 o: A general framework for graph optimization.
\newblock In {\em 2011 IEEE international conference on robotics and automation}, pages 3607--3613. IEEE.

\bibitem[Lau et~al., 2010]{lau2010improved}
Lau, B., Sprunk, C., and Burgard, W. (2010).
\newblock Improved updating of euclidean distance maps and voronoi diagrams.
\newblock In {\em 2010 IEEE/RSJ International Conference on Intelligent Robots and Systems}, pages 281--286. IEEE.

\bibitem[Lau et~al., 2013]{lau2013efficient}
Lau, B., Sprunk, C., and Burgard, W. (2013).
\newblock Efficient grid-based spatial representations for robot navigation in dynamic environments.
\newblock {\em Robotics and Autonomous Systems}, 61(10):1116--1130.

\bibitem[Li et~al., 2021]{li2021optimization}
Li, B., Acarman, T., Zhang, Y., Ouyang, Y., Yaman, C., Kong, Q., Zhong, X., and Peng, X. (2021).
\newblock Optimization-based trajectory planning for autonomous parking with irregularly placed obstacles: A lightweight iterative framework.
\newblock {\em IEEE Transactions on Intelligent Transportation Systems}, 23(8):11970--11981.

\bibitem[Li et~al., 2020]{li2020hrl4in}
Li, C., Xia, F., Martin-Martin, R., and Savarese, S. (2020).
\newblock Hrl4in: Hierarchical reinforcement learning for interactive navigation with mobile manipulators.
\newblock In {\em Conference on Robot Learning}, pages 603--616. PMLR.

\bibitem[Li and et~al., 2019]{li2019neural}
Li, W. and et~al. (2019).
\newblock Teb combined with neural networks for obstacle handling in robotics.
\newblock {\em Artificial Intelligence Review}, 52(4):1873--1888.

\bibitem[Lim et~al., 2021]{lim2021patchwork}
Lim, H., Minho, O., and Myung, H. (2021).
\newblock Patchwork: Concentric zone-based region-wise ground segmentation with ground likelihood estimation using a 3d lidar sensor.
\newblock {\em IEEE Robotics and Automation Letters}.

\bibitem[Lin and Zhang, 2020]{lin2020loam}
Lin, J. and Zhang, F. (2020).
\newblock Loam livox: A fast, robust, high-precision lidar odometry and mapping package for lidars of small fov.
\newblock In {\em 2020 IEEE international conference on robotics and automation (ICRA)}, pages 3126--3131. IEEE.

\bibitem[Liu and Liu, 2023]{10135320}
Liu, C. and Liu, Y. (2023).
\newblock Robot planning and control method based on improved time elastic band algorithm.
\newblock In {\em 2023 4th International Conference on Computer Engineering and Application (ICCEA)}, pages 911--915.

\bibitem[Macenski et~al., 2020]{1729881420910530}
Macenski, S., Tsai, D., and Feinberg, M. (2020).
\newblock Spatio-temporal voxel layer: A view on robot perception for the dynamic world.
\newblock {\em International Journal of Advanced Robotic Systems}, 17(2).

\bibitem[Meng et~al., 2024]{meng2024real}
Meng, D., Chu, H., Tian, M., Gao, B., and Chen, H. (2024).
\newblock Real-time high-precision nonlinear tracking control of autonomous vehicles using fast iterative model predictive control.
\newblock {\em IEEE Transactions on Intelligent Vehicles}.

\bibitem[Nair et~al., 2024]{nair2024predictive}
Nair, S.~H., Lee, H., Joa, E., Wang, Y., Tseng, H.~E., and Borrelli, F. (2024).
\newblock Predictive control for autonomous driving with uncertain, multimodal predictions.
\newblock {\em IEEE Transactions on Control Systems Technology}.

\bibitem[P{\'e}rez et~al., 2024]{perez2024social}
P{\'e}rez, G., Zapata-Cornejo, N., Bustos, P., and N{\'u}{\~n}ez, P. (2024).
\newblock Social elastic band with prediction and anticipation: Enhancing real-time path trajectory optimization for socially aware robot navigation.
\newblock {\em International Journal of Social Robotics}, pages 1--23.

\bibitem[Qie et~al., 2022]{qie2022improved}
Qie, T., Wang, W., Yang, C., Li, Y., Zhang, Y., Liu, W., and Xiang, C. (2022).
\newblock An improved model predictive control-based trajectory planning method for automated driving vehicles under uncertainty environments.
\newblock {\em IEEE Transactions on Intelligent Transportation Systems}, 24(4):3999--4015.

\bibitem[Quinlan, 1995]{quinlan1995real}
Quinlan, S. (1995).
\newblock {\em Real-time modification of collision-free paths}.
\newblock Stanford University.

\bibitem[Rahmani and et~al., 2020]{rahmani2020teb}
Rahmani, K. and et~al. (2020).
\newblock Adaptation of teb for uavs and swarm robotics in complex terrain.
\newblock In {\em 2020 IEEE International Conference on Robotics and Automation}, pages 2341--2346.

\bibitem[Ranganathan, 2004]{ranganathan2004levenberg}
Ranganathan, A. (2004).
\newblock The levenberg-marquardt algorithm.
\newblock {\em Tutoral on LM algorithm}, 11(1):101--110.

\bibitem[R{\"o}smann, 2019]{rosmann2019time}
R{\"o}smann, C. (2019).
\newblock {\em Time-Optimal Nonlinear Model Predictive Control}.
\newblock PhD thesis, Faculty of Electrical Engineering and Information Technology at Technische Universität Dortmund.

\bibitem[R{\"o}smann et~al., 2012]{6309484}
R{\"o}smann, C., Feiten, W., W{\"o}sch, T., Hoffmann, F., and Bertram, T. (2012).
\newblock Trajectory modification considering dynamic constraints of autonomous robots.
\newblock In {\em ROBOTIK 2012; 7th German Conference on Robotics}, pages 1--6. VDE.

\bibitem[R{\"o}smann et~al., 2013]{rosmann2013efficient}
R{\"o}smann, C., Feiten, W., W{\"o}sch, T., Hoffmann, F., and Bertram, T. (2013).
\newblock Efficient trajectory optimization using a sparse model.
\newblock In {\em 2013 European Conference on Mobile Robots}, pages 138--143. IEEE.

\bibitem[R{\"o}smann et~al., 2015]{rosmann2015timed}
R{\"o}smann, C., Hoffmann, F., and Bertram, T. (2015).
\newblock Timed-elastic-bands for time-optimal point-to-point nonlinear model predictive control.
\newblock In {\em 2015 european control conference (ECC)}, pages 3352--3357. IEEE.

\bibitem[R{\"o}smann et~al., 2017a]{rosmann2017integrated}
R{\"o}smann, C., Hoffmann, F., and Bertram, T. (2017a).
\newblock Integrated online trajectory planning and optimization in distinctive topologies.
\newblock {\em Robotics and Autonomous Systems}, 88:142--153.

\bibitem[R{\"o}smann et~al., 2017b]{rosmann2017kinodynamic}
R{\"o}smann, C., Hoffmann, F., and Bertram, T. (2017b).
\newblock Kinodynamic trajectory optimization and control for car-like robots.
\newblock In {\em 2017 IEEE/RSJ International Conference on Intelligent Robots and Systems (IROS)}, pages 5681--5686. IEEE.

\bibitem[R{\"o}smann et~al., 2021]{rosmann2021online}
R{\"o}smann, C., Makarow, A., and Bertram, T. (2021).
\newblock Online motion planning based on nonlinear model predictive control with non-euclidean rotation groups.
\newblock In {\em 2021 European Control Conference (ECC)}, pages 1583--1590. IEEE.

\bibitem[Rsmann et~al., 2021]{9654872}
Rsmann, C., Makarow, A., and Bertram, T. (2021).
\newblock Online motion planning based on nonlinear model predictive control with non-euclidean rotation groups.
\newblock In {\em 2021 European Control Conference (ECC)}, pages 1583--1590.

\bibitem[Rösmann et~al., 2013]{6698833}
Rösmann, C., Feiten, W., Wösch, T., Hoffmann, F., and Bertram, T. (2013).
\newblock Efficient trajectory optimization using a sparse model.
\newblock In {\em 2013 European Conference on Mobile Robots}, pages 138--143.

\bibitem[Rösmann et~al., 2017]{8206458}
Rösmann, C., Hoffmann, F., and Bertram, T. (2017).
\newblock Kinodynamic trajectory optimization and control for car-like robots.
\newblock In {\em 2017 IEEE/RSJ International Conference on Intelligent Robots and Systems (IROS)}, pages 5681--5686.

\bibitem[Shah et~al., 2021]{shah2021rapid}
Shah, D., Eysenbach, B., Kahn, G., Rhinehart, N., and Levine, S. (2021).
\newblock Rapid exploration for open-world navigation with latent goal models.
\newblock {\em arXiv preprint arXiv:2104.05859}.

\bibitem[Shah et~al., 2023]{shah2023vint}
Shah, D., Sridhar, A., Dashora, N., Stachowicz, K., Black, K., Hirose, N., and Levine, S. (2023).
\newblock Vint: A foundation model for visual navigation.
\newblock {\em arXiv preprint arXiv:2306.14846}.

\bibitem[Stefanini et~al., 2024]{stefanini2024efficient}
Stefanini, E., Palmieri, L., Rudenko, A., Hielscher, T., Linder, T., and Pallottino, L. (2024).
\newblock Efficient context-aware model predictive control for human-aware navigation.
\newblock {\em IEEE Robotics and Automation Letters}.

\bibitem[Tan et~al., 2019]{tan2019learning}
Tan, H., Yu, L., and Bansal, M. (2019).
\newblock Learning to navigate unseen environments: Back translation with environmental dropout.
\newblock {\em arXiv preprint arXiv:1904.04195}.

\bibitem[Torrisi et~al., 2016]{torrisi2016variant}
Torrisi, G., Grammatico, S., Smith, R.~S., and Morari, M. (2016).
\newblock A variant to sequential quadratic programming for nonlinear model predictive control.
\newblock In {\em 2016 IEEE 55th Conference on Decision and Control (CDC)}, pages 2814--2819. IEEE.

\bibitem[Van~den Broeck et~al., 2011]{van2011model}
Van~den Broeck, L., Diehl, M., and Swevers, J. (2011).
\newblock A model predictive control approach for time optimal point-to-point motion control.
\newblock {\em Mechatronics}, 21(7):1203--1212.

\bibitem[Wu et~al., 2021]{wu2021improved}
Wu, J., Ma, X., Peng, T., and Wang, H. (2021).
\newblock An improved timed elastic band (teb) algorithm of autonomous ground vehicle (agv) in complex environment.
\newblock {\em Sensors}, 21(24):8312.

\bibitem[Wullt et~al., 2024]{wullt2024model}
Wullt, B., Mattsson, P., Sch{\"o}n, T.~B., and Norrl{\"o}f, M. (2024).
\newblock A model predictive control approach to motion planning in dynamic environments.
\newblock In {\em 2024 European Control Conference (ECC)}, pages 3247--3254. IEEE.

\bibitem[Xiao et~al., 2022]{xiao2022motion}
Xiao, X., Liu, B., Warnell, G., and Stone, P. (2022).
\newblock Motion planning and control for mobile robot navigation using machine learning: a survey.
\newblock {\em Autonomous Robots}, 46(5):569--597.

\bibitem[Yucel et~al., 2021]{yucel2021fuzzy}
Yucel, B., Yilmaz, A., Ervan, O., and Temeltas, H. (2021).
\newblock Fuzzy controlled adaptive follow the gap obstacle avoidance algorithm.
\newblock In {\em Proceedings of the 7th International Conference on Robotics and Artificial Intelligence}, pages 93--98.

\bibitem[Zhou et~al., 2020]{zhou2020autonomous}
Zhou, J., He, R., Wang, Y., Jiang, S., Zhu, Z., Hu, J., Miao, J., and Luo, Q. (2020).
\newblock Autonomous driving trajectory optimization with dual-loop iterative anchoring path smoothing and piecewise-jerk speed optimization.
\newblock {\em IEEE Robotics and Automation Letters}, 6(2):439--446.

\end{thebibliography}

% \section*{Author contributions statement}
% GK Proposed and implemented algorithm. DD and GK were in charge of real-world flight tests and creating a quadrotor for the tests. GK wrote the manuscript in consultation with AK. AK made a critical reviewing the manuscript and final approval for publication.

% \bibliography{jfrExampleRefs}

\end{document}